\documentclass[12pt,authoryear]{elsarticle}

\makeatletter
\def\ps@pprintTitle{%
 \let\@oddhead\@empty
 \let\@evenhead\@empty
 \let\@oddfoot\@empty
 \let\@evenfoot\@empty
 }
\makeatother

\usepackage{amssymb}
\usepackage{amsmath}

\journal{Nuclear Physics B}
\usepackage{natbib}
\usepackage{hyperref}  
\usepackage{subcaption} 
\usepackage{graphicx}
\usepackage{courier}   
\usepackage{longtable}  
\usepackage{array}
\usepackage{xcolor}  
\usepackage{color}      
\usepackage{multicol}  
\usepackage{float} 
\usepackage[utf8]{inputenc}
\DeclareUnicodeCharacter{03BC}{\$mu}
\usepackage[utf8]{inputenc}
\DeclareUnicodeCharacter{FFFD}{ }

\begin{document}

\begin{frontmatter}



\title{AutoFLUKA: A Large Language Model Based Framework for Automating Monte Carlo Simulations in FLUKA  } 


\author[label1]{Zavier Ndum Ndum} 

\affiliation[label1]{organization={Department of Nuclear Engineering, Texas A\&M University},
            country={United States}}

\author[label2]{Jian Tao} 

\affiliation[label2]{organization={College of Performance, Visualization \& Fine Arts, Texas A\&M University}, 
            country={United States}}
\author[label1]{John Ford} 


\author[label1]{Yang Liu\corref{cor1}} 
\cortext[cor1]{Corresponding author. Email: y-liu@tamu.edu} 


\begin{abstract}
Monte Carlo (MC) simulations, particularly using FLUKA, are essential for replicating real-world scenarios across scientific and engineering fields. Despite the robustness and versatility, FLUKA faces significant limitations in automation and integration with external post-processing tools, leading to workflows with a steep learning curve, which are time-consuming and prone to human errors. Traditional methods involving the use of shell and Python scripts, MATLAB, and Microsoft Excel require extensive manual intervention and lack flexibility, adding complexity to evolving scenarios. This study explores the potential of Large Language Models (LLMs) and AI agents to address these limitations. AI agents, integrate natural language processing with autonomous reasoning for decision-making and adaptive planning, making them ideal for automation. We introduce AutoFLUKA, an AI agent application developed using the LangChain Python Framework to automate typical MC simulation workflows in FLUKA. AutoFLUKA can modify FLUKA input files, execute simulations, and efficiently process results for visualization, significantly reducing human labor and error. Our case studies demonstrate that AutoFLUKA can handle both generalized and domain-specific cases, such as Microdosimetry, with an streamlined automated workflow, showcasing its scalability and flexibility. The study also highlights the potential of Retrieval Augmentation Generation (RAG) tools to act as virtual assistants for FLUKA, further improving user experience, time and efficiency. In conclusion, AutoFLUKA represents a significant advancement in automating MC simulation workflows, offering a robust solution to the inherent limitations. This innovation not only saves time and resources but also opens new paradigms for research and development in high energy physics, medical physics, nuclear engineering space and environmental science.
\end{abstract}


\begin{highlights}
    \item AutoFLUKA is an AI-driven tool that automates complex Monte Carlo simulation workflows, integrating seamlessly with FLUKA.
    \item It reduces human intervention and minimizes errors by automating input generation, simulation execution, and post-processing.
    \item AutoFLUKA introduces JSON-based data outputs, enabling easier downstream analysis compared to traditional manual approaches.
    \item AutoFLUKA includes a Retrieval Augmented Generation tool that serves as a virtual assistant, helping users address common FLUKA challenges by providing quick, context-specific guidance.
\end{highlights}


\end{frontmatter}



\section{Introduction}
\label{intro}
Monte Carlo (MC) simulations are virtual experiments that replicate real-world scenarios using statistical methods. These simulations are crucial for accurately modeling complex physical processes, such as radiation transport and particle interactions with matter that could be otherwise expensive, time consuming or resource-intensive. Particularly, FLUKA (FLUktuierende Kaskade) is a versatile MC simulation package with proven applicability across various fields of science and engineering. Originally developed at the European Organization for Nuclear Research (CERN) in the 1960s, then later, jointly by CERN and the Italian National Institute of Nuclear Physics (INFN), FLUKA has evolved into a multi-purpose tool capable of simulating the interaction and transport of over 60 different particles across a wide range of energies in matter \citep{Ahdida2022NewCode, Bohlen2014TheApplications, Fasso2012FLUKA:Package}

 FLUKA simulations have been extensively used to advance research and innovations in nuclear science and engineering. For example, Zhao and colleagues coupled FLUKA with OpenMC  for the physics calculations of higher-energy spallation (p,n) reactions in accelerator-driven sub-critical reactor systems (ADS), which are important for nuclear energy research \citep{Zhao2019ApplicationADS}, while Polanski and colleagues coupled FLUKA with MCNPX to study the production of neutrons in heavy extended targets by electrons of energy from 15 to 1000 MeV \citep{POLANSKI20151}. Bazo and colleagues performed neutron activation analysis, particularly in studying the interaction of neutron beams with different materials using the RP-10 reactor. Here, FLUKA was employed to simulate neutron activation in silicon and germanium \citep{Bazo2018TestingSpectroscopy}. FLUKA has also been used (either independently or coupled with other MC codes like MCNP and GEANT4) to either study or benchmark experimental results  of novel  materials as candidates for nuclear radiation shielding in nuclear power plants, accelerators, medical and radioactive waste management facilities \citep{madbouly2022experimental, el2023effect, singh2021investigations, mansy2022investigation, sasirekha2024investigations, SINGH2021103730, ZAZULA1990385, KUMAR2021103654, AYGUN2020103538}.
 
 In Radiation Therapy and Nuclear Medicine, FLUKA has demonstrated many capabilities. For example, Battistoni and colleagues demonstrated FLUKA's accuracy in simulating depth-dose profiles and lateral dose distributions for therapeutic proton and carbon ion beams, highlighting its crucial role in particle therapy treatment planning \citep{Battistoni2016TheTherapy}. FLUKA has been successfully utilized to develop a dedicated library of PET/CT (positron emission tomography/computed tomography) imaging tools for in vivo verification of treatment delivery and, in particular, PET-based range verification methods for proton therapy. The Code's accuracy for transport of therapeutic ion beams in matter has also been studied \citep{Augusto2018AnTools, Battistoni2015OverviewCode, Botta2013UseImages, Ortega2013AFLUKA, Nasirzadeh2022ModelingCode, Sommerer2009In-beamTargets, Sommerer2006InvestigatingMatter}. FLUKA is the standard simulation package for radiation protection studies at CERN, with G. Battistoni and colleagues discussing its applications in studying radiation damage to electronics at high-energy hadron accelerators \citep{Battistoni2011FLUKAAccelerators}. 2011). FLUKA has also been valuable in studying and optimizing radiation detectors for both space and therapeutic applications. For instance, it has been used to characterize the response of an innovative active neutron spectrometer called DIAMON   against neutron fields at high and low energies \citep{Braccini2022AAccelerators}. Its Capabilities for Microdosimetric Analysis (i.e, its ability to simulate energy deposition spectra in a tissue-equivalent proportional counter (TEPC) and produce a reliable estimate of delta-ray events when such a TEPC is exposed to high-energy heavy ions (HZE)) was investigated by Northum \citep{Northum2012FLUKAAnalysis}. As a result, it was used to design and characterize the response of  a TEPC  to GRCs and also on the surface of Mars \citep{Northum2015SimulatedMars, Beck2005FLUKARadiation}, while Bortot and colleagues also utilized FLUKA simulations to design and benchmark the experimental response of a novel Avalanche-confinement TEPC against different therapeutic beams \citep{Bortot2017ALevel, Bortot2018AFields, Bortot2018MonteBeam, Bortot2020ACNAO, Mazzucconi2019AProcessing}.  

It is evident from above descriptions that FLUKA’s comprehensive physics models, continuous development and active user forum has made it a valuable tool for advancing research and development in fields like High-energy physics, Medical Physics, Space Science, Nuclear engineering and Environmental science \citep{Ahdida2020MeasurementsAntiprotons, Ahdida2022NewCode, Iliopoulou2018MeasurementsCSBF}. The code is available as a package for Linux and UNIX platforms and for Windows via the Windows Subsystem for Linux (WSL). It comes with a graphical user interface (GUI) called FLAIR offering capabilities for input file creation (through input cards), geometry visualization, simulation execution and results visualization. Both programs can either be downloaded from the official FLUKA website\footnote{\href{http://www.fluka.org/fluka.php}{The official FLUKA site: FLUKA home} }  or CERN\footnote{\href{https://fluka.cern/}{Home | The official CERN FLUKA website} }    and require a FLUKA license to use. 

Despite being a robust simulation tool, FLUKA’s inherent limitations in automation and integration with external post-processing tools pose significant challenges to users. A typical simulation workflow from start to finish is represented in Figure 1. Each step involves a chain of operations, some degree of reasoning and decision making and large sets of data to be processed. Traditionally, shell scripts and Python scripts have been employed to reduce human labour related to input file preparation, simulation execution, and data extraction, while software like MATLAB and Microsoft Excel have been employed to post-process simulation data. However, these approaches often require extensive manual intervention, are prone to errors, and lack flexibility when adapting to complex and evolving simulation scenarios.
\begin{figure}
    \centering
    \includegraphics[width=1\linewidth]{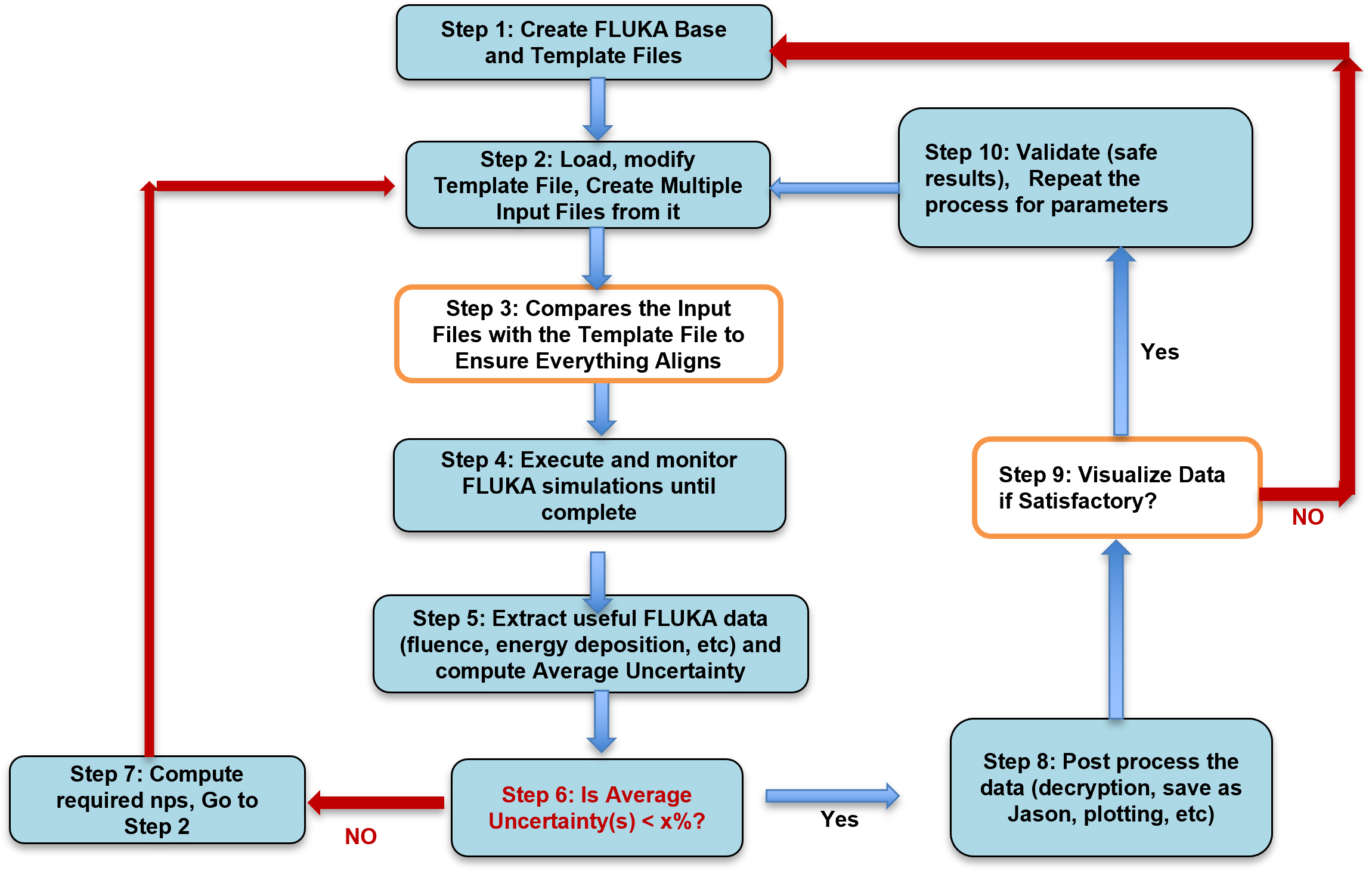}
    \caption{Schematic of a hypothetical FLUKA workflow adopted for the automation. Steps 3 and 9 are highlighted because they require a human in the loop to verify the FLUKA-Fortran code syntax and to judge the accuracy of the results respectively.}
    \label{fig1:fluka_workflow}
\end{figure}

AI and machine learning have made significant progress in nuclear engineering, including data-driven modeling \citep{liu2018data, liu2022data,liu2022sam, yaseen2023fast}, uncertainty quantification \citep{lin2021uncertainty, liu2021uncertainty, liu2019uncertainty, wu2018inverse, liu2019validation}, reactor optimization\cite{radaideh2021physics}, reactor transient analysis\citep{prantikos2023physics}, and digital twins \citep{lin2022development}. These advancements have provided new methodologies to tackle complex problems, improve predictive capabilities, and enhance decision-making processes in nuclear systems. 

More recently, large language models (LLMs) such as ChatGPT have sparked significant public interest in the use of AI for general purposes. A recent example is Kwon's use of LLMs to analyze public sentiment towards nuclear energy \citep{kwon2024sentiment}. More importantly, AI agents powered by LLMs have emerged as powerful systems for engineering applications by combining natural language processing with autonomous reasoning for decision-making and adaptive planning \citep{Reid2024GeminiContext, Jin2024FromFuture, Zhang2024AdvancingStudies, Zhang2024ADiscovery}. These agents can learn and adjust their behaviour based on inputs/outputs from their environment, making them suitable for automation in science and engineering \citep{Murthy2024REX:AGENTS}. Notable examples include AutoGPT \citep{Yang2023Auto-GPTHe}, BabyAGI \citep{YoheiN.B.2023GitHubYoheinakajima/babyagi}, WebGPT \citep{Nakano2021WebGPT:Feedback}, OpenAGI \citep{Ge2023OpenAGI:Experts}, MetaGPT \cite{Hong2023MetaGPT:Framework}, HuggingGPT \citep{Shen2024Hugginggpt:Face} Generative agents \citep{Park2023GenerativeBehavior}, etc which all automate various tasks, like web search, content creation, business and finance tasks, and many more. There also exist some domain-specific applications of AI Agents like Data Interpreter, for solving data-centric scientific problems  \citep{Hong2024DataScience}, HxLLM-a framework that utilizes LLMs to aid in the design and optimization of Heat Exchangers \citep{Mishra2024DesignModels} and MechAgents, for automatically solving mechanics problems through a multi-agent approach \citep{Ni2024MechAgents:Knowledge}. In Nuclear Science and Engineering, researchers have assessed ChatGPT's strengths and weaknesses in the design and safety analysis of nuclear reactors. They recommended prompt engineering, fine-tuning, and Retrieval Augmented Generation (RAG) to address the LLM’s weaknesses like social bias, hallucinations and vulnerability to adversarial prompts \citep{Athe2023TestingApplications}. 

However, LLMs and AI Agents applied to the Monte Carlo Simulations of Radiation Interaction with matter have remained largely understudied. It is in this light that we propose, AutoFLUKA , an AI agent-powered application developed to automate typical Monte Carlo (MC) simulation workflows performed in FLUKA. The primary motivation behind this innovation stems from challenges encountered while using FLUKA to design and optimize a TEPC for radiation characterization in Microdosimetry. Not only did we develop an automation scheme to address this domain-specific problem, but we also extended our application to the general example described in the FLUKA manual \citep{Ferrari2024Fluka:2024} so that users across different disciplines could benefit. We demonstrated that through a set of carefully crafted instructions, otherwise called prompts and a single click, AutoFLUKA can be instructed to modify the contents of a valid FLUKA input file, execute the simulation and efficiently process the simulation results for visualization, thus saving the user a tremendous amount of time, labour while mitigating the possibility of human errors. AutoFLUKA was developed using the LangChain Python Framework\citep{H.Chase2023GitHubApplications}, with most of its components currently compatible with both gpt-4o from OpenAI and gemini 1.5 from Google DeepMind, thus making it versatile for both paid and free LLM API keys. 

Also, AutoFKUKA is developed to integrate seamlessly with FLUKA and to be; (i)- \textbf{scalable} by implementing a tool approach, thus giving users the possibility to customize and/or add tools for post-processing of data according to their needs and; (ii)-\textbf{flexible} through a step-wise prompting technique, meaning that it can be prompted to repeat or start the workflow from any step. For instance, if the simulations already ran on another computer,  we could import them and ask  AutoFKUKA to search and decrypt the binary files and plot the spectra for visualization. Moreover, it is important to state that the methods employed in this paper for FLUKA can be applied to any simulation software with a text-based input file system like the Monte Carlo N-Particle\footnote{\href{https://mcnp.lanl.gov/}{mcnp.lanl.gov}}  (MCNP) code, Particle and Heavy Ion Transport code System\footnote{\href{https://phits.jaea.go.a/}{phits.jaea.go.a}}  (PHITS), etc. This opens up new research paradigms in Nuclear Science and Engineering.  

The rest of the paper is organized in the following manner: Section 2 covers a brief overview of LLMs, AI Agents and the LangChain Python framework. Section 3 provides  a detailed description of the building blocks of AutoFLUKA; Section 4 presents AutoFLUKA on two case studies for a generalized case and a domain-specific Microdosimetry case respectively, including a Retrieval Augmentation Generation (RAG) tool which intends to act as a FLUKA virtual assistant. Section 5 finally presents concluding remarks and future working directions.

\section{MATERIALS AND METHODS}
\label{sec2}

\subsection{From LLMs to AI Agents: Using LLMs Programmatically}
\label{subsec2.1}

Large language models (LLMs) can be defined as a group of artificial intelligence (AI) or transformer neural networks containing a substantial number of parameters, often reaching tens or hundreds of billions \citep{Zhao2023AModels}. The higher the number of parameters, the better the performance. These models are trained on extensive corpora, enabling them to efficiently perform a wide variety of natural language processing (NLP) tasks like text generation, summarizing, translation, general coding, question answering and even acting as chatbots. 

Since the stunning debut of ChatGPT in November 2022, several other significant AI models and tools have been released like Gemini from Google DeepMind (2024), LLaMA from Meta (2023), Claude by Anthropic (2023), Copilot by Microsoft, etc. This rapid development has seen profound effects across industry, academia and science through increased automation and efficiency, new research paradigms and accelerated discovery. However, because these models are trained on general datasets, they often suffer from hallucinations, misinformation or increased bias and fairness inherited from the training data. Even though techniques whereby, the LLM chatbot is prompted to think critically using keywords or phrases like “think step by step”, “ take a deep breath”, etc (referred to as the Chain of thought (CoT)), or given one or a few examples of the task before asking it to perform similar tasks (referred to as one-shot/few-shots) have been used to overcome the above limitations, the latter still poses privacy concerns when working with personal or proprietary data. 

\begin{figure}
    \centering
    \includegraphics[width=0.9\linewidth]{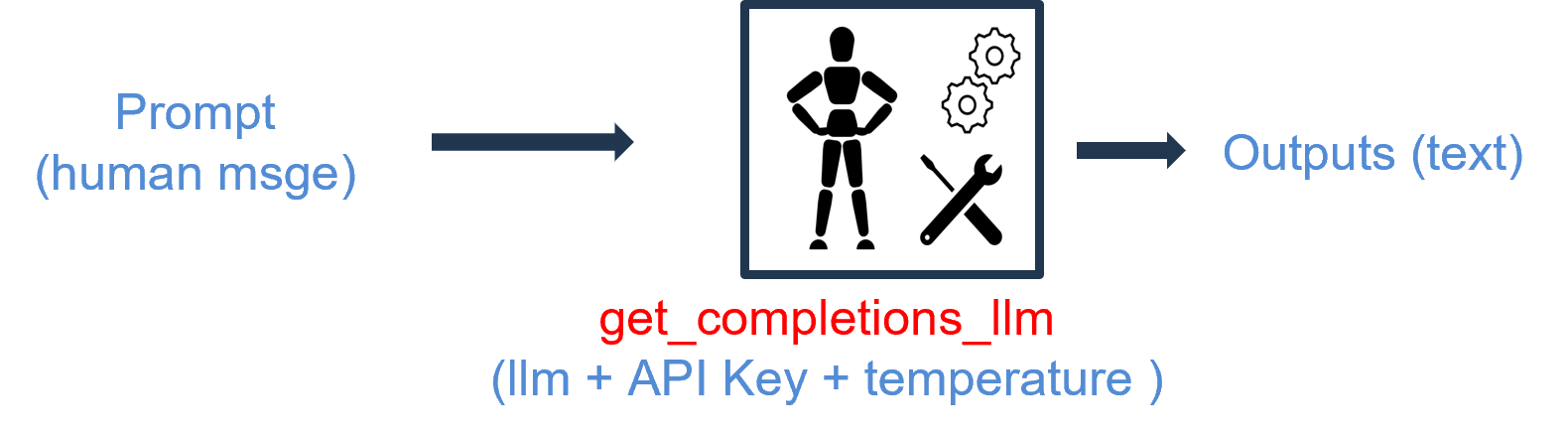}
    \caption{A simple chat completion model. important parameters include specifying the LLM type  with the  corresponding API key  and temperature}
    \label{fig:simple_llm}
\end{figure}

To address the above concerns, LLMs can rather be used programmatically, thanks to application programming interfaces (APIs). Through this, they can then be safely augmented and/or fine-tuned with domain-specific and proprietary data for in-context learning and completion of specific (in-house) tasks that are otherwise not possible or restricted with generalized chatbots. The basic syntax involves initializing the LLM model (gpt-4o, gemini-1.5 flash, etc) with the appropriate API key and the temperature as shown in Figure \ref{fig:simple_llm}. The temperature here represents the degree of randomness of the model’s responses. A low value (close to 0) means deterministic (less random or precise) responses, while a high value (close to 1 and above) leads to more creative (surprising) responses. Once set, the model then uses its built-in tools (general knowledge) to respond to user queries.

To demonstrate this, a simple chat model called \textbf{get\_completions\_llm}  was coded from scratch, initialized using the OpenAI API key, temperature = 0.3 and prompted with tasks of varying complexity. Table 1 shows some responses from the model. Through this, we found that this model excelled in completing simple tasks but for more complicated ones like file reading/editing, and code execution, it struggled. 

\begin{figure}
    \centering
    \includegraphics[width=1\linewidth]{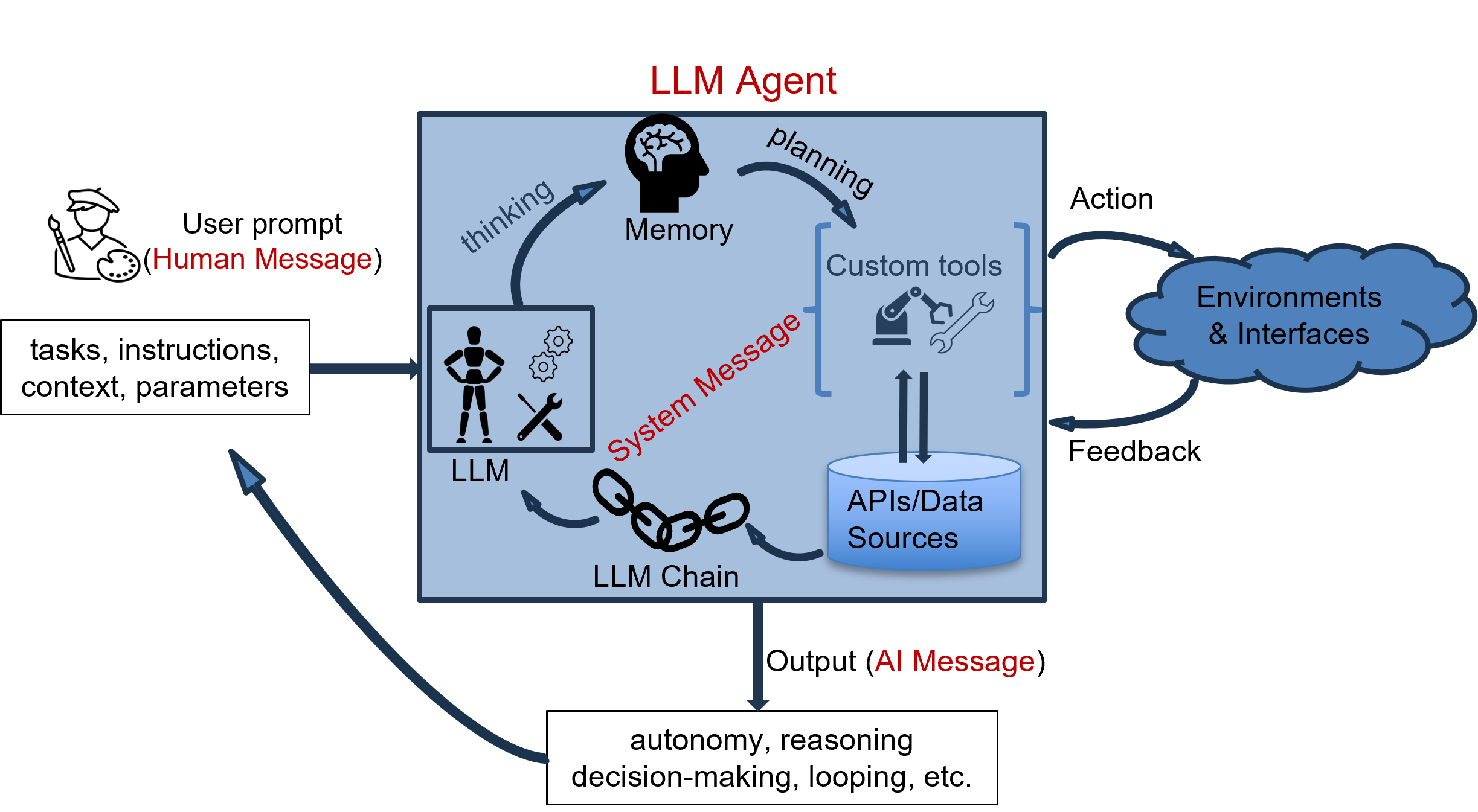}
    \caption{Schematic of an LLM Agent workflow. With additional tools and prompt fine-tuning, the agent can utilize the thought capabilities of LLMs to perform actions by connecting with interfaces, invoking and executing tools on external environments while receiving feedback and optimizing the process until success.}
    \label{fig:llm_agent}
\end{figure}

Accomplishing such complicated tasks and many others requires careful crafting of the prompts and the integration of additional components or tools into the core LLM to activate memory capabilities, thought processes and interaction with external environments and interfaces. \textit{This leads to the so-called LLM Chains and LLM agents (or AI Agents) as represented in  Figure \ref{fig:llm_agent} and whose development and deployment are the subject of this research as will be discussed in detail in the proceeding sections}.

\begin{scriptsize} 
\begin{longtable}{|p{3.5cm}|p{9.5cm}|}
\caption{Responses from a simple text completion model "get\_completions\_llm" programmatically initialized with the OpenAI API key, "llm = gpt-4" and temperature = 0.3. This model excels at simple tasks but struggles with complex ones.} \\
\hline
\textbf{User query (Human message)} & \textbf{Response from \ttfamily get\_completions\_llm \normalfont } \\ \hline

\textbf{User query}: What is the answer to 4*6+7-8/9? Be brief but show the steps \newline
\textbf{Category}: Simple 
& 
\ttfamily
To solve the expression (4 * 6 + 7 - $\frac{8}{9}$), follow the order of operations (PEMDAS/BODMAS): \newline
1. \textbf{Multiplication and Division} (from left to right): [ 4 * 6 = 24 ] [ $\frac{8}{9} \approx 0.8889$ ] \newline
2. \textbf{Addition and Subtraction} (from left to right): [ 24 + 7 = 31 ] [ 31 - 0.8889 $\approx$ 30.1111 ] \newline
So, the answer is approximately \textbf{30.1111.} 
\normalfont
\newline
\textbf{Remark}: Successful \\ \hline

\textbf{User query}: What is FLUKA in 10 words? \newline
\textbf{Category}: Simple 
& 
\ttfamily
FLUKA is a particle physics simulation software for radiation transport.
\normalfont
\newline
\textbf{Remark}: Successful \\ \hline

\textbf{User query}: Please execute this Python function: \{file\_reader\} \newline
\textbf{Category}: Complex 
& 
\ttfamily
I'm unable to execute code directly. However, I can help you understand how to execute a Python function and provide guidance on how to run it. \newline
To execute a Python function, you typically need to: \newline
1. Define the function in your script or import it if it's defined in another module. \newline
2. Call the function with the appropriate arguments. \newline
\textbf{… Continues} 
\normalfont
\newline
\textbf{Remark}: Not Successful \\ \hline

\textbf{User query}: file\_path = 'fluka\_example.inp' \newline
prompt: Please, execute the FLUKA input code located in this directory \{file\_path\}. \newline
\textbf{Category}: Complex 
& 
\ttfamily
I'm unable to execute files or run code directly. However, I can help you understand how to run a FLUKA input file on your system. Here are the general steps you would follow: \newline
1. \textbf{Install FLUKA}: Ensure that you have FLUKA installed on your system. You can download it from ... \newline
2. \textbf{Prepare Your Environment}: Set up your environment correctly, ... \newline
3. \textbf{Run the Input File}: \newline
    - Open a terminal or command prompt. \newline
    -...\textbf{… Continues} 
\normalfont
\newline
\textbf{Remark}: Not Successful \\ \hline

\end{longtable}
\end{scriptsize}

\subsection{Building Agentic Applications on LangChain}
\label{subsec2.2}

LangChain is an open-source framework designed for developing applications powered by LLMs. It was introduced by Harrison Chase in 2023 \citep{H.Chase2023GitHubApplications}, aiming to make the integration of LLMs into AI-powered applications more accessible, especially for tasks requiring complex, scalable workflows or multi-step reasoning. 

LangChain was chosen for this project because of (i)-its wide acceptance in the AI community with a proven history of applications build to automate tasks spanning from customer service \citep{Pandya2023AutomatingOrganizations}, architecture \citep{Jeong2023GenerativeFramework} to health care\citep{Singh2024RevolutionizingModel} and; (ii)-its versatility and a rich library of components and methods like; \textbf{Schemas} which refer to the data structures (text, ChatMessages and documents) used across LLM platforms;  \textbf{LLM Wrappers} for connection to LLM models like those from OpenAI, Google DeepMind and Hugging Face; \textbf{Prompt Templates} for structuring and optimizing input prompts, thus reducing the labor involved in hard-coding text for the LLM's understanding and interpretation; \textbf{Indexes} for efficient retrieval of relevant information from user queries; LLM Chains for linking multiple LLM components together; \textbf{Agents} and \textbf{Agent executors} for executing custom tools and connecting to external environments via APIs. 
As represented in the schematic in Figure 2(b), the agent, after receiving user input or query (Human Message) exploits the reasoning and thinking capabilities of its core LLMs model(s) and LMM chains which contain well-crafted prompt templates  (System Messages instructing the LLM how to complete tasks) to plan and invoke appropriate tools to interact with the external world, perform actions, while receiving feedback until the task is marked as Completed or Finished and the output (AI Message) returned to the user.

For demonstration purposes, a simple agent named \textbf{simple\_llm\_agent} was constructed using the LangChain Framework. A simple Python script to read and parse file contents (function, named \textbf{file reader}) was also written and passed to the agent as a tool. We also added a \textbf{Wikipedia search tool} for interacting with the internet and the \textbf{Python REPL tool} for code generation and deployment. This was to verify the Agent’s ability to select the right tool from the mix to perform a particular task. Results from querying the agent are shown in Table 2. The input consisted of the \textbf{System Message} (instructions given to the agent on how to accomplish tasks) and the \textbf{Human Message} which is the user query or prompt consisting of the action(s) or task(s) to the accomplished. The output (\textbf{AI message}) also consists of the right tool invocation, the required action and a request for further assistance as shown.

\begin{scriptsize}   
\begin{longtable}{|p{2cm}|p{9cm}|}
\caption{simple\_llm\_agent's input and output showing the invocation and execution of a tool to read and print file contents} \\

\hline
\multicolumn{2}{|c|}{\textbf{INPUT}} \\ \hline
\textbf{System Message} & 
You're an expert. Where necessary, invoke the necessary tools with the correct parameters to complete tasks. If you do not have the answer, DO NOT try to make up one. \\ \hline
\textbf{Human (user query)} & 
Please read the file contents located in this directory: \texttt{tepc\_template\_file.inp} and print the first 10 lines. \\ \hline

\multicolumn{2}{|c|}{\textbf{OUTPUT}} \\ \hline
\textbf{TOOL} & TOOL: file\_reader\_tool$\rightarrow$call\_5UZiF2Y9SZIKnVBg3neYN9Ng \\ \hline

\textbf{Rendered Output} & 
Here are the first 10 lines of the file \texttt{tepc\_template\_file.inp}: \newline
\ttfamily
1. TITLE \newline
2. TEPC GEOMETRY: DESIGN AND OPTIMIZATION \newline
3. *Set the defaults for precision simulations \newline
4. DEFAULTS \hspace{1cm} HADROTHE \newline
5. *Define the beam characteristics \newline
6. ------------------------------------------- \newline
7. *Thermal Neutron Field \newline
8. ------------------------------------------- \newline
9. *Delta x/y >= 25 mm (Catania Beam width) \newline
10. @scale=2000 \newline
\normalfont
If you need more information or further assistance, please let me know! \\ \hline

\textbf{Original File (first 16 lines)} & 
\ttfamily
1. TITLE \newline
2. TEPC GEOMETRY: DESIGN AND OPTIMIZATION \newline
3. *Set the defaults for precision simulations \newline
4. DEFAULTS \hspace{1cm} HADROTHE \newline
5. *Define the beam characteristics \newline
6. ------------------------------------------- \newline
7. *Thermal Neutron Field \newline
8. ------------------------------------------- \newline
9. *Delta x/y >= 25 mm (Catania Beam width) \newline
10. @scale=2000 \newline
11. BEAM \hspace{0.3cm} -0.062 \hspace{0.2cm} 12 \hspace{0.2cm} -2.5 \hspace{0.3cm} 2.5 \hspace{0.2cm} HEAVYION \newline
12. HI-PROPE \hspace{0.5cm} 6 \hspace{0.5cm} 12 \newline
13. *Define the beam position \newline
14. *w.r.t Geo\_axis \newline
15. ------------------------------------------- \newline
16. BEAMPOS \hspace{0.5cm} -110 \hspace{0.5cm} 0.0 \hspace{0.5cm} 0.93 \hspace{0.5cm} 0.0 \newline
... 
\normalfont
\\ \hline
\end{longtable}
\end{scriptsize}  

Note that the \textbf{file\_reader\_tool} was deliberately designed to read and parse the entire file content (we have shown just 16 lines here for brevity). However, as illustrated in Figure 2(b), the AI agent (\textbf{simple\_llm\_agent}) in this case was able to observe the tool’s output, reason, plan and then act by printing out just the first 10 lines of the file content as instructed. \textit{This \textbf{Train of Thought (ToT)} approach can also be used to edit the file content with specific user-defined parameters in which case the\textbf{ REPL tool} will be invoked. It is one of the key strategies exploited in this research to develop \textbf{AutoFLUKA’s} suite of tools as discussed further in the next sections.}  

\section{AutoFLUKA’s CUSTOM TOOLS }
\label{sec3}

It is important to note that each block of the FLUKA simulation workflow shown in Figure \ref{fig1:fluka_workflow} involves a series of sub-tasks. For instance, creating the FLUKA input files requires loading the template file, modifying it with the user-defined parameters before making multiple versions with varying seeds (statistically independent cycles). For this reason, different custom tools were developed, each addressing just a single block in most cases. Table \ref{table3: AutoFLUKA tools} briefly summarizes these tools for both the generalized example from the FLUKA manual \citep{Ferrari2024Fluka:2024} and the domain-specific case from a paper describing TEPC design and optimization in Microdosimetry \citep{Ndum2024DesignApplications}. The sections that follow will describe the main sections both simulation workflows and how the corresponding tools were crafted to accomplish the task. 

\begin{table}[h!]
\centering
\caption{FUKAAGENT’s Tool Library}
\label{table3: AutoFLUKA tools}
\begin{scriptsize}  
\begin{tabular}{|p{4cm}|p{8cm}|}
\hline
\textbf{Tool Name} & \textbf{Description} \\ \hline

\multicolumn{2}{|l|}{\textbf{Generalized Example Workflow}} \\ \hline
python\_repl\_tool & Generates and deploys useful Python code \\ \hline
csv\_file\_reader\_tool & Reads a CSV file and returns its content as a dictionary \\ \hline
text\_file\_reader\_tool & Reads and returns the contents of a text file \\ \hline
fluka\_input\_file\_creator\_tool & Modifies and creates multiple FLUKA input files from the template \\ \hline
fluka\_executer\_tool & Executes FLUKA on the command line via subprocess \\ \hline
fluka\_data\_decrypter\_tool & Decrypts the binary output file “\_fort.xx”, xx being any number from 17 through 99 \\ \hline
fluka\_data\_to\_json\_tool & Stores useful data as JSON for downstream processing \\ \hline
nps\_and\_uncertainty\_tool & Extracts the 'Average Uncertainty' and Total primaries Run or Number of particles simulated (NPS) \\ \hline
fluka\_data\_plotter\_tool & Plots useful FLUKA data \\ \hline
fluka\_assistant\_tool & RAG tool to help the user with FLUKA challenges \\ \hline

\multicolumn{2}{|l|}{\textbf{Microdosimetry Workflow (All the above, plus these extra tools)}} \\ \hline
weight\_data\_with\_gas\_gains\_tool & Weights the spectral counts (energy deposited in the sensitive volume) with TEPC gas gains if needed \\ \hline
lin\_to\_log\_rebinning\_tool & Converts the raw data (spectral counts) from a linear to a logarithmic binning \\ \hline
microdosimetric\_spectra\_tool & Computes the Microdosimetric spectra and quality factor, plots the results and saves them as JPEG images. Also saves the lineal energy distributions as a CSV file. \\ \hline

\end{tabular}
\end{scriptsize}  
\end{table}

\subsection{Creating the Template File. }
\label{subsec3.1}

The template file must be manually created from a valid FLUKA input file; valid here means that the input file must be running correctly and producing the expected results. FLUKA uses a text-based input file system. These input files are structured with specific keywords related to the so-called input cards which are used to configure different aspects of the simulation related to the geometry, sources, materials, physics or particle generation and transport and scoring (like energy deposition, dose, particle fluence, current, yield, etc). A version of this file will be made in our Git Up repository together with the template. Our template version has been configured to allow the user to externally modify parameters through variables without interfering with the FLUKA code. These variables are stored in a CSV file and read by the \textbf{csv\_file\_reader\_tool}. 

Also important to note is that Since FLUKA is written in Fortran, ALL  input files must be column-based to respect Fortran’s fixed-column syntax. As such a Python formatting configuration has been adopted to ensure that parameter values always align at that right column 21 no matter their length. Tools like the \textbf{text\_file\_reader\_tool} and the \textbf{fluka\_input\_file\_creator\_tool} were all tailored to accomplish this task. 

\subsection{Executing FLUKA and Decrypting the Generated Binary Files}
\label{subsec3.2}

Once the input files have been created, the \textbf{fluka\_executer\_tool} then lunches and monitors the simulations, capturing the standard input and outputs (stdin/stdout) and any errors generated by FLUKA itself (stderr). A bash script written in SLURM (\textbf{Simple Linux Utility for Resource Management}) for scheduling and submitting the jobs on the command line was incorporated within this tool to be executed using Python’s subprocess. 

In FLUKA, the user can select any Fortran logical unit “xx” between 21 and 99 (17 by default for the DETECT card) for their detector(s) which can also be any scoring card as shown in Table \ref{table4}.  FLUKA then generates results for every cycle in binary format, automatically following a naming convention. For example, if the base file was named  “example.inp”,  and the user selected the logical unit (BIN 46) to score the boundary crossing fluence using the USRBDX card, then the generated results from running multiple cycles will appear as \textbf{example\_01001\_fort.46, example\_02001\_fort.46}, etc. Even though FLAIR provides a straightforward way to decrypt and process or merge these data, users running FLUKA through the command line and on clusters with limited access to FLAIR must do this manually for each cycle using the appropriate FLUKA’s in-built post-processing utilities as shown in Table 4. Though these utilities work in an interactive pattern, the user must type respective file names for all cycles in the simulation and then specify a file name in which the output (decrypted data) will be stored. This process, requiring a lot of writing is time-consuming, monotonous and could be subject to typographical errors. 

Through our custom data decrypting tool (\textbf{fluka\_data\_decrypter\_tool}), this entire process has been automated. Since we only know which utility works for the various cards  as presented in Table 4, BUT cannot determine which card was employed by the user or what quantity was scored from the binary file’s name alone, our decryption tool employs a trial-by-error approach to determine the appropriate utility. During this decryption process, the stdin, stdout and stderr which by default print on the console are captured and saved in a text file named \textbf{“decryption\_logs”} to reduce token consumption by the LLM Agent.  To ensure continuity in naming convention, the resulting outputs; that is the merged (sum over the total number of cycles simulated or  \textbf{“\_sum.lis”})  and the tabulated data (\textbf{“\_tab.lis}) are also named similarly to include the Fortran binary identifier, xx (e.g. \textbf{“\_fort\_xx\_sum.lis”} and \textbf{“\_fort\_xx\_tab.lis”}). 

Summarily, the \textbf{“\_sum.lis”} file contains integrated data over all histories or events in the simulation. It conatains quantities like fluence, energy deposition, or dose in specific regions, averaged over the NPS, together with uncertainties calculated using standard deviations to give a sense of the statistical accuracy of the simulation. It is useful for users interested in just an overview of the overall simulation results. On the other hand, the \textbf{“\_tab.lis”} file stores the same data in a bin-wise tabulated format. That is the differential and double differential distributions in terms of energy bins, angular bins, spatial bins, or time bins, depending on what was scored in the simulation. It is useful for users who want to understand how the output varies across bins.  Note that only a brief description has been provided here and the reader is advised to consult the FLUKA manual \citep{Ferrari2024Fluka:2024} for a detailed description of these files.

\begin{scriptsize}
\begin{longtable}{|p{1.5cm}|p{6.3cm}|p{2cm}|p{1.5cm}|}
\caption{FLUKA Post-processing utilities for decrypting binary output into human-readable data. The logical unit identifier can be any number “xx” between 21 and 99 (17 by default for the DETECT card).} 
\label{table4}
\\
\hline
\textbf{Utility} & \textbf{Function/Description} & \textbf{Scoring Card} & \textbf{Logical Unit “xx”} \\ \hline
\endfirsthead

\hline
\textbf{Utility} & \textbf{Function/Description} & \textbf{Scoring Card} & \textbf{Logical Unit “xx”} \\ \hline
\endhead

usruvw.f & For decrypting the residual nuclei output data & RESNUCLEi &  21–99 \\ \hline
usryvw.f & To read out the particle yield output data & USRYIELD & 21–99 \\ \hline
usxsuw.f & Processes the boundary crossing fluence, current, etc data & USRBDX & 21–99 \\ \hline
usbsuw.f & For processing the spatial distribution of particle fluence, energy deposition, dose, etc & USRBIN & 21–99 \\ \hline
ustsuw.f & For the track length estimator data & USRTRACK &  21–99 \\ \hline
detsuw.f & To sum energy deposition results & DETECT & 21–99 \\ \hline

\end{longtable}
\end{scriptsize}  

The \textbf{fluka\_data\_to\_json\_tool} extracts useful data from these two file types and stores them in Jason format in a file named \textbf{fluka\_data.json}. This \textbf{fluka\_data.json} file is structured in a nested and hierarchical form, containing various data entries and lists as shown in Appendix B. The average uncertainty was also computed by this \textbf{fluka\_data\_to\_json\_tool} from the differential distributions and reported in the \textbf{“\_tab.lis”} sections using the following formula.

\begin{equation}
\hat{U} = \frac{\sum_{i=1}^{N_{\text{bins}}} c_i \cdot U_i}{\sum_{i=1}^{N_{\text{bins}}} c_i}
\end{equation}

\noindent where, $c_i$ and $U_i$ are respectively the counts and error (\%) of each bin interval.

This value, which decreases with the number of contributions (NPS or $N$) as $1/\sqrt{N}$, was invoked by the AI Agent during its workflow using the \texttt{nps\_and\_uncertainty\_tool} to determine the statistical accuracy of the simulations through the following relation.

\begin{equation}
\hat{U} \sim \frac{1}{\sqrt{N}} \quad \Rightarrow \quad N_2 > \left( \frac{\hat{U}_1}{\hat{U}_2} \right)^2 N_1
\end{equation}

We also incorporated a visualization tool (\textbf{fluka\_data\_plotter\_tool}) to plot and store the data as JPEG images in the same working directory for visualization.

The \textbf{fluka\_assistant\_tool} was developed based on the Retrieval Augmented Generation (RAG) approach for LLMs. This tool, augmented with extra knowledge from the FLUKA user manual and a few discussions from the FLUKA help forum has already demonstrated remarkable capabilities in addressing domain-specific questions that are otherwise complex for generalized chatbots like ChatGPT, Gemini, etc. It is therefore envisaged that with proper augmentation and fine-tuning, this tool will evolve into a virtual assistant (NOT to replace FLUKA experts) but to address simple, straightforward questions within the FLUKA User Forum, before directing users to the experts for more complicated problems. This will save FLUKA experts, a considerable amount of time to focus on FLUKA code augmentation and fine-tuning of some of the existing physics models for particle interaction and transport in matter or adding new ones. 
	
Finally, \textbf{weight\_data\_with\_gas\_gains\_tool}, \textbf{lin\_to\_log\_rebinning\_tool} and  \textbf{microdosimetric\_spectra\_tool} are additional tools that have been incorporated to be used for the domain-specific case of detector design and optimization for radiation characterization in Microdosimetry. Their descriptions have been provided in Table \ref{table3}.	

\subsection{Single Agent and Multi-Agent Workflows}
\label{subsec3.3}

In this project, both the single and multi-agent approaches were exploited to automate the simulation workflow. For the single-agent approach, the \textbf{simple\_llm\_agent} model developed earlier in section 2.1 was augmented by adding more tools, fine-tuning the prompt templates (system messages) and giving it memory capabilities for retaining the history of previous runs (particularly useful for the FLUKA Assistant RAG tool). The Agent Executor in this case acts both as a supervisor, selecting the appropriate tool for each task and monitoring the workflow until all steps have been completed. On the other hand, the multi-agent approach incorporated each custom tool with its own LLM agent having a specific system message. All the agent nodes were then connected using the LangGraph method from LangChain. In this approach, the actions of these individual agents (also known as workers) are coordinated by a supervisor, which, itself is an agent with a specific system message. Figure 3(a) illustrates these differences, while Figure 3(b) shows a graph representation of the multi-agent framework.

\begin{figure}[htbp]
    \centering
    \begin{subfigure}[!t]{0.8\linewidth} 
        \centering
        \includegraphics[width=\linewidth]{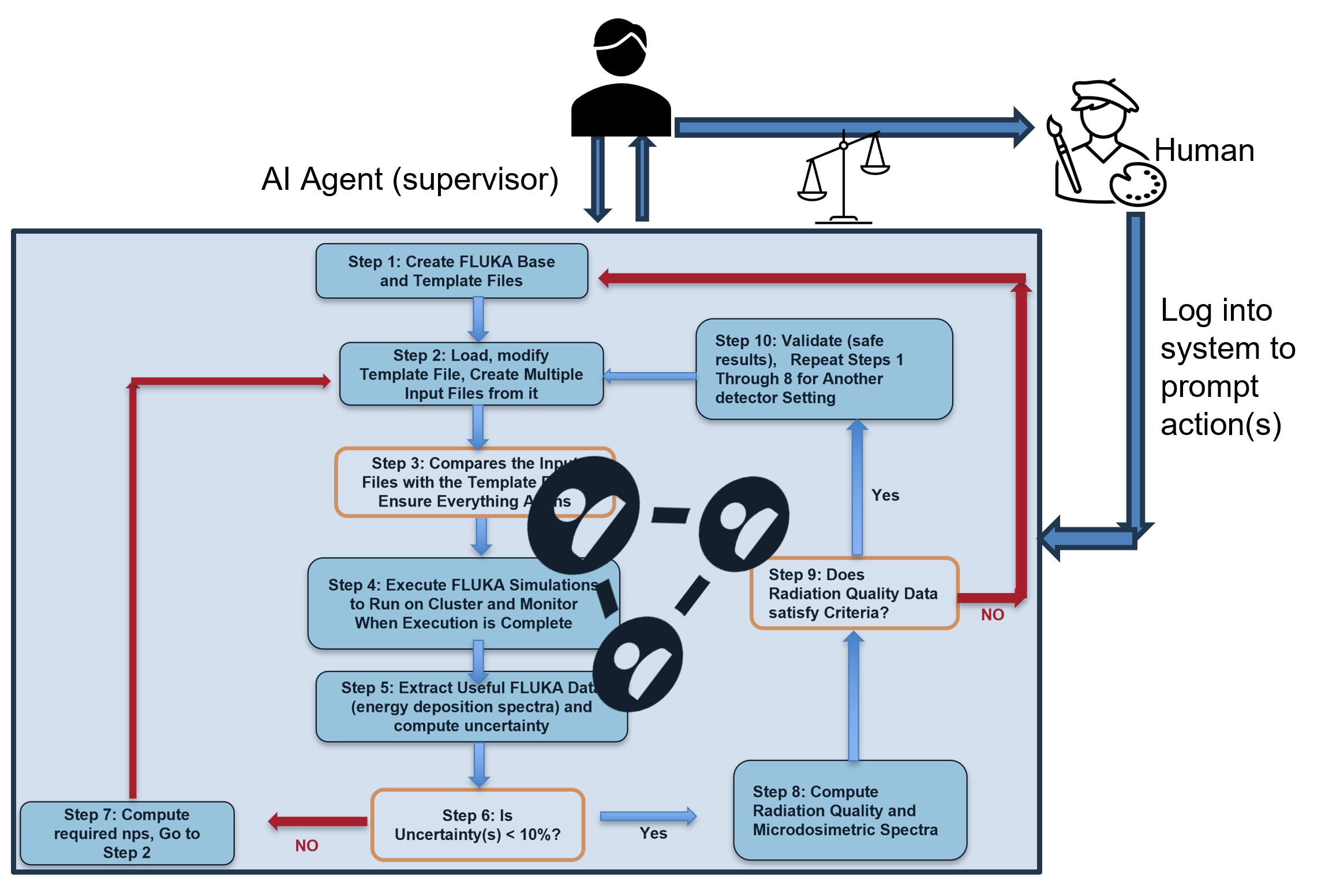}
        \caption{}
        \label{fig4:first-image}
    \end{subfigure}
    
    \vspace{0.1cm} 

    \begin{subfigure}[!t]{0.7\linewidth} 
        \centering
        \includegraphics[width=\linewidth]{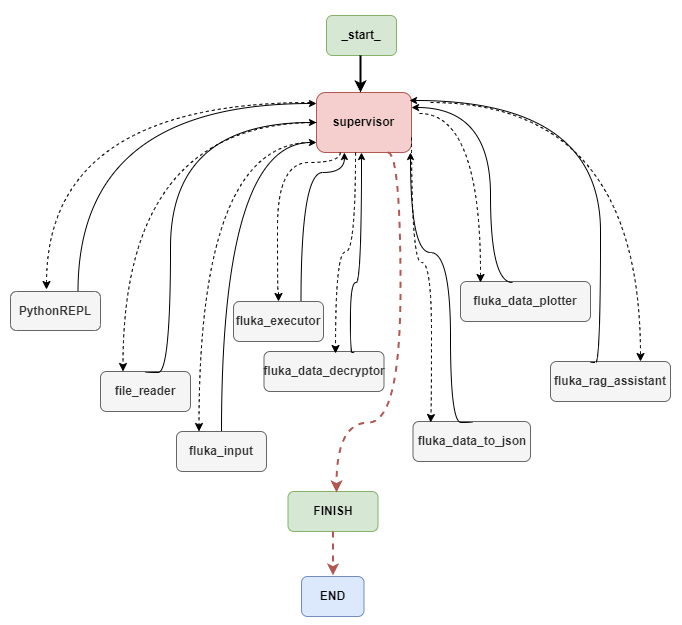}
        \caption{}
        \label{fig4:second-image}
    \end{subfigure}
    
    \caption{Schematic view of the multi-agent workflow, showing the supervisor AI agent at the top, coordinating actions of different agents within the blocks.  (b) –Graph visualization of the multi-agent workflow. Notice that each agent takes actions from as well as reports back to the agent supervisor until the task is marked as complete, after which the FINISH + END sequence is triggered. The human in the loop is to log into the system to initiate the action.}
    \label{fig:two-images}
\end{figure}

A comparison between these two revealed that despite the inherent flexibility of the multi-agent approach, in that each agent or worker can be fine-tuned with a detailed (long) prompt template (system message and description) on how to accomplish tasks, it constantly hallucinated. Here, the agent supervisor either calls the wrong worker for a particular action (what we call worker-action mismatch) or tries to invoke a non-existent worker. Detailed crafting of the user query and exclusion of passive tools (workers not directly involved in the task) were applied to mitigate this effect. On the other hand, the single-agent approach, because of its inherent simplicity saw less hallucinations. The major limitation being the context length of tool descriptions which currently is 128,000 tokens for gpt-4o.  

The human in the loop currently has two functions: (i)- logs into the computer system and initiates the action. This is the sole responsibility of the human for security reasons, and (ii)-checks the accuracy of the simulations, from ensuring that the input files are created correctly (step 3) to evaluating the generated plots (step 9). Step 3 is quite tricky since it involves ensuring that the generated input files respect Fortran’s column-based syntax as described earlier in section 3.1. This process is still time-consuming. Therefore, investigating how this could be achieved automatically through the so-called few-shot and chain-of-thought approaches, especially with the more flexible multi-agent approach is currently under study. 

\section{RESULTS AND DISCUSSIONS}
\label{sec4}

In this section, we present the results from applying AutoFLUKA on two used cases; the general example treated in the FLUKA manual (Ferrari et al., 2024) and the domain-specific case of detector development in Microdosimetry based on this ongoing research \citep{Ndum2024DesignApplications}. These results were obtained from the Single-Agent framework. 

\subsection{AutoFLUKA on the General Example }
\label{subsec4.1}

This FLUKA example scores the pion fluence inside and around a proton-irradiated Beryllium (Be) target. The geometry of the setup is shown in Figure 4. It consists of a mono-energetic proton beam of momentum 50 GeV/c (gigaelectronvolts per unit of the speed of light) at 50 cm from two slabs of Be, labelled regBe3 and regBe4 standing for region 3 and region 4 respectively. The entire system is placed in a vacuum (region 2 or regVa2 as shown). The beam impinges from the left and crosses to the right, while interacting and generating secondary particles (pions) within each region. The boundary crossing fluence and current from region 3 (regBe3) to region 4 (regBe4), the pion fluence in and around the target and other quantities are scored as represented in Table \ref{table5}.  In this case, AutoFLUKA was asked to create five (05) input files corresponding to five statistically independent input cycles (i.e., formatted with the same user-defined parameters) but with different seeds.  The complete execution sequence of the workflow is reported in Tables \ref{table6}, \ref{table7}, \ref{table8} and Figure \ref{fig:dose-distr}. 

\begin{figure}
    \centering
    \includegraphics[width=0.8\linewidth]{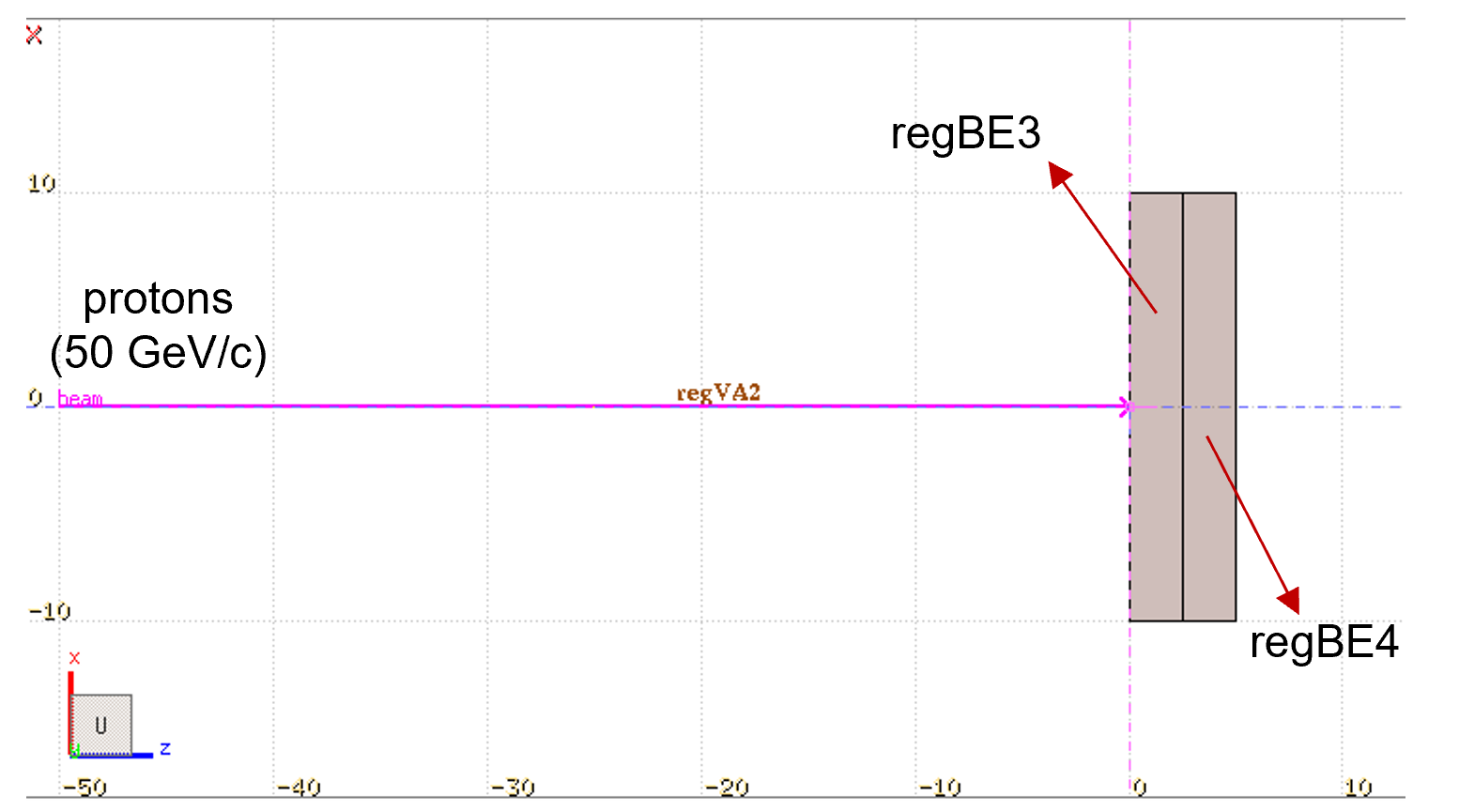}
    \caption{FLUKA geometry of the setup for the “Charged pion fluence inside and around a proton-irradiated Beryllium (Be) target”. A proton beam of momentum = 50 GeV/c impinges on a slap of Be from the left and crosses through the right slap. Secondary particles like pions are generated in these two slaps and their corresponding fluence and current scored.  }
    \label{fig:enter-label}
\end{figure}

First, as shown in Table \ref{table6}, the Input is made up of two parts; (i)-“System” or System Message which has been hard-coded into the agent’s initialization scheme using LangChain’s “ChatPromptTemplate” method; and (ii)- “Human” or Human Message (user query) which in this case is a prompt containing the instructions given to AutoFLUKA to complete the task. It is important to note that certain keywords and action phrases can significantly alter the agent’s behaviour, thus, careful crafting using prompt engineering techniques was done to arrive at this current working version of the prompt. A stepwise approach with keys like complete file directories and file names was implemented to reduce hallucinations. Also, through this, AutoFLUKA can easily be prompted to go back and repeat certain steps in the workflow depending on its current observation of the output from other tools.

\begin{small}  
\begin{longtable}{|p{8.5cm}|p{2cm}|p{1.2cm}|}
\caption{Scored quantities from the Charged pion fluence inside and around a proton-irradiated Beryllium (Be) target example.} 
\label{table5}
\\
\hline
\textbf{Quantity Scored} & \textbf{Scoring Card} & \textbf{Logical unit} \\ \hline
\endfirsthead

\hline
\textbf{Quantity Scored} & \textbf{FLUKA Card} & \textbf{Logical unit} \\ \hline
\endhead

Boundary crossing fluence from regBE3 to regBE4 & USRBDX & 46 \\ \hline
Boundary crossing current from regBE3 to regBE4 & USRBDX & 47 \\ \hline
Track length fluence inside target (Upstream) & USRTRACK & 48 \\ \hline
Track length fluence inside target (Downstream) & USRTRACK & 49 \\ \hline
Pion fluence inside and around the target & USRBIN & 50 \\ \hline
Energy Deposition inside the target & USRBIN & 51 \\ \hline
\end{longtable}
\end{small}

Table\ref{table7} shows the first part of the output from the workflow (steps 1 to 4). The CSV file reader tool reads the parameters file and returns its contents as a dictionary (a total of 21 parameters were used here just for demonstration purposes). It then reports back to the supervisor which invokes the \textbf{fluka\_input\_file\_creator\_tool} to modify the template file and create statistically independent input files. Next, the \textbf{fluka\_executor\_tool} is called to execute the shell script that launches the simulations on the command line. It then monitors and records the start/end time and calculates the total simulation time. Then, the data is decrypted and stored in one place as a Jason file, while extracting the calculated average uncertainty and NPS (\textbf{nps\_and\_uncertainty\_tool}). 

\begin{scriptsize}
\begin{longtable}{|p{1.3cm}|p{11.4cm}|}
\caption{AutoFLUKA’s input (system message and user query). The system message was hardcoded into the agent using LangChain’s ChatPromptTemplate method, while this user query is the prompt.}
\label{table6}
\\
\hline
\multicolumn{2}{|c|}{\textbf{INPUT}} \\ \hline

\textbf{System Message} & 
\textbf{You're an expert. You have the ability to write and execute FLUKA code and process output data.} 
Where necessary, invoke the tools you have been given with the correct parameters to complete tasks. If you do not have the answer, \textbf{DO NOT try to make up one.} \\ \hline

\textbf{User Query \newline (Human Msge)} & 
Please, execute the following workflow in a chronological order from steps 1 through 8: \newline
\textbf{step 1}: Read and modify the contents of \texttt{C:/Users/.../example\_template.inp} with parameters from this csv file: \texttt{C:/Users/.../parameters.csv}. \newline
\textbf{step 2}: Create 5 FLUKA input copies with prefix 'example' in the directory \texttt{C:/Users/.../Results}. \newline
\textbf{step 3}: When all the input files have been created, execute FLUKA, \textbf{AND Decrypt all data} from output files ending in [\_fort.xx], where [xx] is any number between 17 through 99. \newline
\textbf{step 4}: Next, process the resulting data from output files ending in [\_sum.lis] and [tab.lis] and save the results as \texttt{fluka\_data.json}. Extract the \textbf{[Average Uncertainty]} from the [output\_fort\_46\_tab.lis] tab\_section \textbf{AND} the [Total Primaries Run] or [output\_fort\_46\_sum.lis] sum\_section of the first detector from this \texttt{fluka\_data.json} data. \newline
\textbf{step 5}: \textbf{REMEMBER:} If this Average Uncertainty is less than 10\%, then GO TO Step 8. Else, GO TO Step 6. \newline
\textbf{step 6}: \textbf{\textcolor{blue}{Given that the Average Uncertainty is inversely proportional to the square root of the nps, what value of nps (rounded to the next 100000th) is needed to get an uncertainty less than 10\%?}} \newline
\textbf{step 7}: Next, \textbf{UPDATE the CSV} file \texttt{C:/Users/.../parameters.csv} with this new nps, then REPEAT steps 1 to 4, SKIP STEP 5 and CONTINUE from step 8. \newline
\textbf{step 8}: Next, plot the data from the resulting \texttt{fluka\_data.json} file with the following specifications: \newline
\texttt{plot blocks=False, log scale=False, semilogx=True, semilogy=False}. SAVE the results as 'jpeg' images in the same directory. When done with the workflow, go to \textbf{FINISH}. \\ \hline
\end{longtable}
\end{scriptsize}  

These first steps are just serial, prompting the agent to select the appropriate tool, while steps 5 through 8 push AutoFLUKA to think and act, thus showcasing the reasoning and decision-making capabilities of LLMs and AI agents. Once the calculated average uncertainty does not meet the user’s requirements in step 5, a thought process is performed (step 6) to determine the required number of primaries (NPS). It then invokes the Python REPL tool that generates and deploys Python code to compute this value and correctly updates the CSV file as instructed. Once this is done, the workflow is repeated from step 1 as instructed in step 7.

\begin{scriptsize}
\begin{longtable}{|p{2cm}|p{11cm}|}
\caption{Out from the workflow showing the various tool invocations; the user-defined parameters from a CSV file that were used to modify the input file (21 in this case); the last input file (example\_05.inp) is created and written in the directory as shown; the shell script executions of FLUKA and total simulation time is recorded; the total Primaries Run (nps) and Average Uncertainty is also extracted which will be used by the Agent to decide the statistical accuracy of the simulation and decide the next step.}
\label{table7}
\\

\hline
\multicolumn{2}{|c|}{\textbf{OUTPUT}} \\ \hline

\textbf{CSV file reader (returns data as dictionary)} & 
\ttfamily \textbf{TOOL:csv\_file\_reader\_tool$\rightarrow$call\_iycGF3F6G6ldFonLs37TVbUQ} 
\begin{multicols}{2}  
beam\_pos\_z: "-50" \newline
emfcut\_E\_kin: "0.01" \newline
emfcut\_E\_gamma: "0.01" \newline
usrbdx\_Emin: "1e-9" \newline
usrbdx\_Emax: "50" \newline
usrbdx\_Nbins: "100" \newline
usrtrack\_Emin: "0.001" \newline
usrtrack\_Emax: "50" \newline
usrtrack\_yol: "1000" \newline
usrtrack\_bins: "20" \newline
\columnbreak  
usrbnin\_Xmin: "-50" \newline
usrbnin\_Xmax: "50" \newline
usrbnin\_Ymin: "-50" \newline
usrbnin\_Ymax: "50" \newline
usrbnin\_Zmin: "-50" \newline
usrbnin\_Zmax: "50" \newline
usrbnin\_Nx: "100" \newline
usrbnin\_Ny: "100" \newline
usrbnin\_Nz: "60" \newline
seed: "10" \newline
nps: "600000" 
\end{multicols}
\normalfont \\ \hline

\textbf{Create input files} & 
\ttfamily \textbf{TOOL:fluka\_input\_file\_creator\_tool$\rightarrow$call\_Ovtv3iIGKcg5T7SvplBInDy} \newline
Formatted files written to \newline
C:/Users/.../FLUKA\_example/Results/example\_05.inp 
\normalfont \\ \hline

\textbf{Execute FLUKA simulations} & 
\ttfamily \textbf{TOOL:fluka\_executer\_tool$\rightarrow$call\_Z6dJXJ19ipE7BFpK4EFMea} \newline
Execution Directory: /mnt/c/Users/.../Results \newline
How Many Input files? 5 \newline
Now Running: \newline 
AutoFLUKA\_job1.sh for : example\_01.inp \newline
AutoFLUKA\_job2.sh for : example\_02.inp \newline
AutoFLUKA\_job3.sh for : example\_03.inp \newline
AutoFLUKA\_job4.sh for : example\_04.inp \newline
AutoFLUKA\_job5.sh for : example\_05.inp \newline
All Jobs are running... \newline
Simulations successfully executed at: 2024-09-09 13:11:06 \newline
Simulation Time: 00:00:34.652984 
\normalfont \\ \hline

\textbf{Binary to human-readable} & 
\ttfamily \textbf{TOOL:fluka\_data\_decrypter\_tool$\rightarrow$call\_yn0E4Y4MvFXbI81mk511CJ7t} \newline
FLUKA Binary Files Decryption complete.
\normalfont \\ \hline

\textbf{Organize data in JSON format} & 
\ttfamily \textbf{TOOL:fluka\_data\_to\_json\_tool$\rightarrow$call\_meIo3lgAvFj9gb8CA51yJft} \newline
fluka\_data.json file created 
\normalfont \\ \hline

\textbf{Extract uncertainty and NPS} & 
\ttfamily \textbf{TOOL:nps\_and\_uncertainty\_tool$\rightarrow$call\_QdkcqvDNVUO1Mmhv9eKURdxhh} \newline
\textbf{Average Uncertainty: 42.7305060397987 \newline
Total Primaries Run: 3000000}
\normalfont \\ \hline

\end{longtable}
\end{scriptsize}  

At this point, we prompted FKUKAAGENT to skip the uncertainty evaluation step to avoid an endless cycle since the NPS calculation is only approximate and might still result in an uncertainty value slightly greater than the required threshold. This is because as clearly stated in the FLUKA training courses and further confirmed though private discussions in User Forum, the uncertainty (variance) is itself a radom variable, meaning that runs with the same NPS but different random seeds will give different values of the variance. Also, by rounding up the calculated NPS, we avoided running the workflow multiple times for a single case study thus, saving simulation time. Table 8 shows this second run with the new simulation time. After this, the output is decrypted as before, and the data extracted an saved ot the \textbf{fluka\_data.json} file as before. Then, the visualization tool is invoked to plot the data and save the plots as images in the same directory. 

\begin{scriptsize}  
\begin{longtable}{|p{1.4cm}|p{11.5cm}|}
\caption{Since the average uncertainty did not meet the requirement specified in the prompt (<10\%) the Agent invoked the Python REPL tool, created and deployed code that calculated the required NPS and updated the parameters. A second run of simulations with this new NPS took about 1 hour, 16 minutes and 29 seconds. The visualization tool then plotted the data according to the specifications in the prompt and the AI Agent summarized the steps performed.}
\label{table8}
\\

\hline
\multicolumn{2}{|c|}{\textbf{OUTPUT}} \\ \hline

\textbf{Computes required NPS} & 
\ttfamily \textbf{TOOL: python\_repl\_tool$\rightarrow$call\_ReyIvOcreagvH5youJQRqngf} \newline
Successfully executed: \newline
\newline
``python \newline
import math \newline
\newline
\# Given values \newline
current\_uncertainty = 42.7305060397987 \newline
current\_nps = 3000000 \newline
\newline
target\_uncertainty = 10 \# target is less than 10\% \newline
\newline
\# Uncertainty is inversely proportional to the square root of nps \newline
\# (current\_uncertainty / target\_uncertainty) = sqrt(current\_nps / target\_nps) \newline
\newline
\# Rearranging to solve for target nps \newline
scaling\_factor = (current\_uncertainty / target\_uncertainty) ** 2 \newline
target\_nps = current\_nps * scaling\_factor \newline
\newline
\# Round to the next 100000th \newline
rounded\_target\_nps = math.ceil(target\_nps / 100000) * 100000 \newline
rounded\_target\_nps \newline
...\newline
\normalfont \\ \hline
\textbf{AI Message} &
The calculated value of nps needed to achieve an uncertainty of less than 10\% is 55,000,000. Next, I will update the CSV file with this new nps value and repeat steps 1 to 4, skipping step 5. \newline
\textbf{TOOL: python\_repl\_tool$\rightarrow$call\_d7KK58FJ1lilionlPz01N0t}
\normalfont \\ \hline

\textbf{Updates CSV parameters file} & 
\ttfamily Successfully executed: \newline
``python \newline
import csv \newline
\newline
\# File paths \newline
csv\_file\_path = 'C:/Users/.../parameters.csv' \newline
\# New nps value \newline
new\_nps = 55000000 \newline
\newline
\# Read the CSV file \newline
with open(csv\_file\_path, mode='r') as file: \newline
\hspace{0.5cm} reader = csv.DictReader(file) \newline
\hspace{0.5cm} data = list(reader) \newline
\newline
\# Update the nps value \newline
for row in data: \newline
\hspace{0.5cm} row['nps'] = new\_nps \newline
\# Write the updated data back to the CSV file \newline
with open(csv\_file\_path, mode='w', newline='') as file: \newline
\hspace{0.5cm} fieldnames = data[0].keys() \newline
\hspace{0.5cm} writer = csv.DictWriter(file, fieldnames=fieldnames) \newline
\hspace{0.5cm} writer.writeheader() \newline
\hspace{0.5cm} writer.writerows(data) \newline
\# Confirm the update \newline
with open(csv\_file\_path, mode='r') as file: \newline
\hspace{0.5cm} updated\_data = list(csv.DictReader(file)) \newline
updated\_data \newline
... 
\normalfont \\ \hline

\textbf{Execute FLUKA simulations} & 
\ttfamily \textbf{TOOL: fluka\_executer\_tool$\rightarrow$ call\_LcoiG0VNrXpt5upfE2pXlPD} \newline
Execution Directory: /mnt/c/Users/.../Results \newline
How Many Input files? 5 \newline
Now Running: \newline
AutoFLUKA\_job1.sh for : example\_01.inp \newline
AutoFLUKA\_job2.sh for : example\_02.inp \newline
AutoFLUKA\_job3.sh for : example\_03.inp \newline
AutoFLUKA\_job4.sh for : example\_04.inp \newline
AutoFLUKA\_job5.sh for : example\_05.inp \newline
All Jobs are running... \newline
Simulations successfully executed at: 2024-09-09 14:28:25 \newline
Simulation Time: 1:16:29.729428 
\normalfont \\ \hline

\textbf{Binary to JSON } & 
Same as above; same function calls but with different call IDs.
\normalfont \\ \hline

\textbf{Plot Data for Visualization} & 
\ttfamily \textbf{TOOL: fluka\_data\_plotter\_tool$\rightarrow$call\_zcs6qiOKpN4Muhm20ely7Tg} \newline
json\_file: C:/Users/.../fluka\_data.json \newline
plot\_error\_bars: false \newline
plot\_blocks: false \newline
semilogx: true \newline
output\_dir: C:/Users/.../FLUKA\_example/Results 
\normalfont \\ \hline

\textbf{AI-generated Summary of Steps} & 
\ttfamily \textbf{AI: Rendered Output} \newline
\newline
The workflow has been successfully completed. The plots have been generated and saved as specified. \newline
\newline
\textbf{Summary of Steps:} \newline
1. \textbf{Read and Modify Template:} The template file was read and modified with parameters from the CSV file. \newline
2. \textbf{Create Input Copies:} Five FLUKA input copies were created with the prefix 'example'. \newline
3.\textbf{ Execute FLUKA and Decrypt Data:} FLUKA simulations were executed, and the resulting data files were decrypted. \newline
4. \textbf{Process Data:} The resulting data was processed and saved as fluka\_data.json. The average uncertainty and total primaries run were extracted. \newline
5. \textbf{Check Uncertainty:} The average uncertainty was found to be greater than 10\%, so the required nps to achieve an uncertainty less than 10\% was calculated. \newline
6. \textbf{Update CSV and Repeat:} The CSV file was updated with the new nps value, and steps 1 to 4 were repeated, skipping step 5. \newline
7. \textbf{Plot Data:} The data from the resulting fluka\_data.json file was plotted with the specified settings and saved as JPEG images. \newline
\normalfont
\textbf{Next Step:} \newline
- Review the generated plots and the fluka\_data.json file for further analysis. \newline
- If you have any further tasks or need additional assistance, please let me know! 
\normalfont \\ \hline

\end{longtable}
\end{scriptsize}  

Even though not prompted to, a detailed summary of the workflow was also generated by the agent, asking the user to check the generated plots and the Jason file, showcasing the reasoning capabilities of LLMs. These generated plots are shown in Figure 5. In the FLUKA manual(Ferrari et al., 2024), it is mentioned that the \textbf{“\_tab.lis”} data is just a bin-wise representation of the spectral and cumulative quantities from the \textbf{“\_sum.lis”} file. This was confirmed by the plots in  Figure 5(b and d). The rest of the plots from the \textbf{“\_fort\_47”} to \textbf{“\_fort\_49”} files have been omitted for brevity. The Jason file, which encapsulates the detailed hierarchical data entries and lists as described in section 3.2 is shown in Appendix B.

\begin{figure}[H]
    \centering
    \begin{subfigure}[!t]{0.32\linewidth} 
        \centering
        \includegraphics[width=\linewidth]{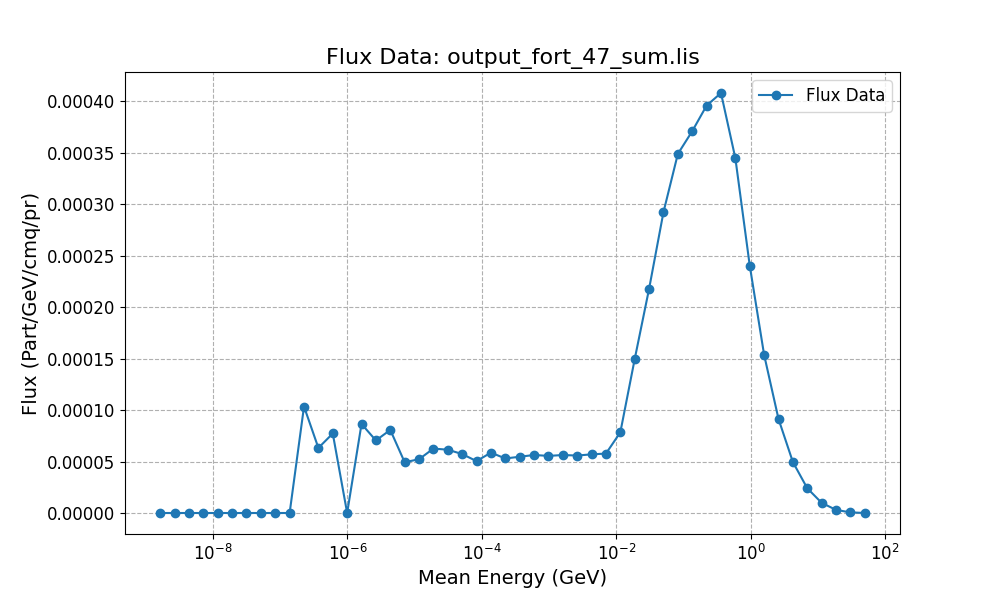}
        \caption{}
        \label{fig6:first-image}
    \end{subfigure}
    \hfill
    \begin{subfigure}[!t]{0.32\linewidth} 
        \centering
        \includegraphics[width=\linewidth]{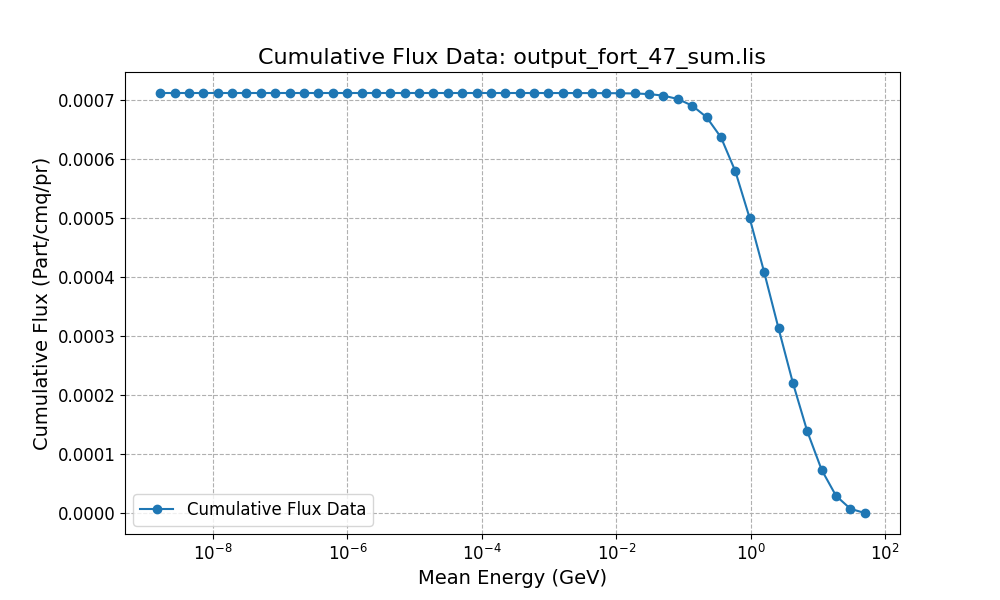}
        \caption{}
        \label{fig6:second-image}
    \end{subfigure}
    \hfill
    \begin{subfigure}[!t]{0.32\linewidth} 
        \centering
        \includegraphics[width=\linewidth]{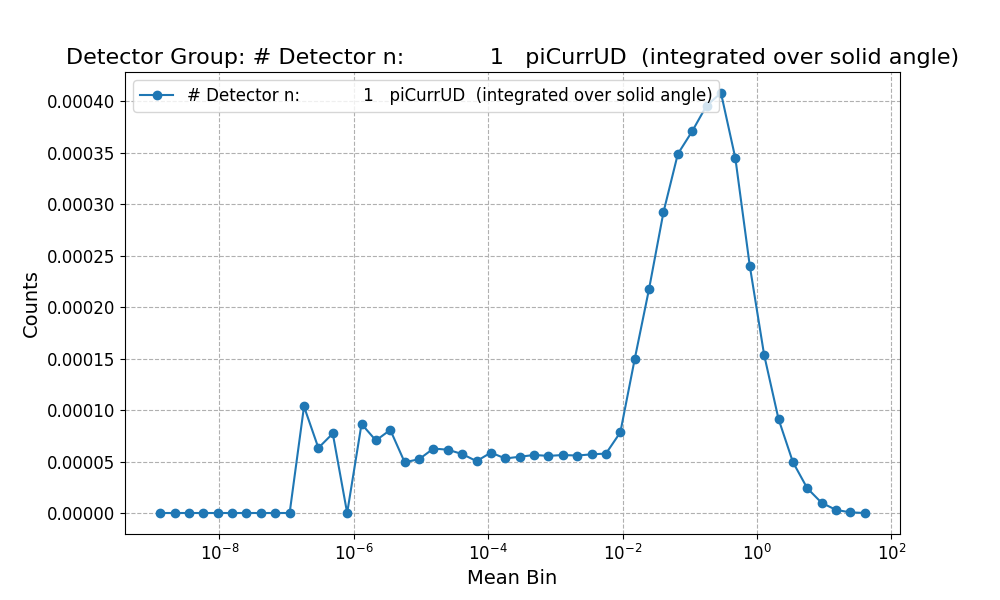}
        \caption{}
        \label{fig6:third-image}
    \end{subfigure}
    
    \caption{Plots generated during the workflow; (a)- Spectral flux extracted from the \textbf{“\_sum.lis”} output file; (b)- cumulative flux also extracted from the \textbf{“\_sum.lis”} output file; (c)- Spectral flux extracted from the bin-wise \textbf{“\_tab.lis”} output file. Notice that this is identical to the \textbf{“\_sum.lis”} file data and was done for verification purposes.}
    \label{fig6:three-images}
\end{figure}

It is envisaged that from this general demonstration, users will be able to adapt the workflow to their domain-specific cases. It is important to state that running FLUKA, just like any other Monte Carlo simulation software (MCNP, PHITS,  GEANT4, etc) is not trivial and often involves a learning curve. The availability of training courses and a vibrant FLUKA User Forum contributes tremendously to addressing user problems ranging from code development and runtime errors to data processing. However, it usually takes considerable time (from days to weeks) to get these queries completely addressed. This is why we incorporated \textbf{fluka\_assistant} within our suite of tools, NOT to replace the FLUKA experts but to serve as a “middleman”  or virtual assistant when fully augmented with domain knowledge, thus helping users to develop and deploy code faster as will be presented in the next section.

\subsection{AutoFLUKA’s RAG Tool as a Virtual Assistant}
\label{subsec4.2}

In this section, we demonstrate AutoFLUKA’s ability (using our inbuilt RAG tool) to respond correctly to domain-specific questions, thus helping the FLUKA user to resolve errors within their code (input file) and to correctly post-process the simulation results.  This domain-specific case is the “detector design and optimization for radiation quality measurements in Microdosimtry”, which will also be presented in the next section. 

The RAG tool developed here is scalable and can take multiple PDFs. Specifically designed to handle scientific documents with mathematical expressions, this tool combines the PyMuPDF and PyPDFLoader Libraries to extract text from documents. It also leverages Optical Character Recognition (OCR) via the PyTesseract library to extract text from images that might be contained in the PDFs. Then, the Regular Expression Library (regex or re) is extensively used to identify and capture mathematical notations like LaTeX-style inline and block equations. After this, embeddings are created from the extracted text using Chroma DB, a highly efficient vector store for storing and querying document embeddings as shown in Figure 6. Embeddings are vector representations of the document content.

The RAG tool is currently compatible with OpenAI and Gemini, meaning that embeddings can be generated and queried using either OpenAI or Google Gemini (depending on which LLM model has been initialized by the user through the API). During the embedding process, the text is split into smaller chunks (e.g., paragraphs, sections, etc). To ensure that the chunking retains logical coherence, preventing the loss of context, overlapping between chunks is enabled. These embeddings are then stored persistently, ensuring that the computationally expensive process is performed only once. When new PDF files are added to the directory, the system detects them, processes only the new content, and updates the existing vector store, ensuring that redundant re-embedding of previously processed documents is avoided. Chroma DB allows for efficient querying by transforming user queries into vectors performing a similarity search to retrieve the most relevant document vectors, which are then used to generate explicit responses by the LLM. The memory capabilities of AutoFLUKA (i.e. the ability to recall the history of previous queries and responses) enable a continuous interaction with the tool’s knowledge base by asking follow-up questions as shown in Appendix --

For our domain-specific case, the RAG tool was augmented with knowledge from the FLUKA manual (Ferrari et al., 2024). Due to FLUKA licensing constraints, we only added our own discussions from the user forum on uncertainty propagation and DETECT card spectra. We also included  a Logbook of ideas gathered during the design phase of this detector (all in PDF format). 

\begin{figure}[htbp]
    \centering
    \begin{subfigure}[!t]{0.48\linewidth} 
        \centering
        \includegraphics[width=\linewidth]{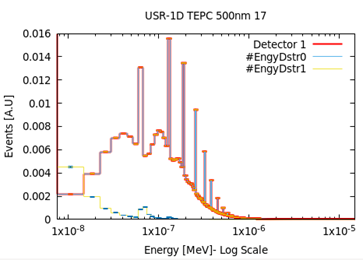}
        \caption{}
        \label{Figure_7_a.png}
    \end{subfigure}
    \hfill
    \begin{subfigure}[!t]{0.48\linewidth} 
        \centering
        \includegraphics[width=\linewidth]{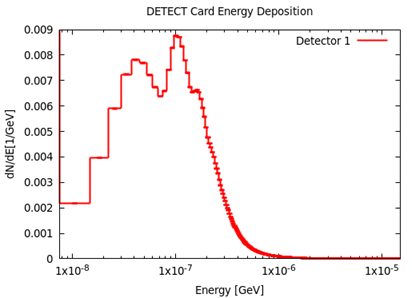}
        \caption{}
        \label{Figure_7_b.png}
    \end{subfigure}
    
    \caption{(a) -Energy deposition spectrum in a low-density Tissue Equivalent gas recorded by DETECT card in FLUKA showing unwanted spikes due to wrong settings in the physics cards; (b) The same code with the inclusion of EMFCUT (PROD-CUT activation), DELTARAY, MULSOPT (with single scattering activation) and removal of PART-THRES and EMFFIX cards as recommended yielded the correct results.  AutoFLUKA was able to reproduce these recommendations even more concisely}
    \label{fig7:two-images}
\end{figure}


\subsubsection{Spikes in the DETECT Card Spectra}
\label{subsubsection4.2.1}
This error resulted during the early stages of the detector development research due to wrong settings of the Physics (production and transport) cards. After exchanges with the FLUKA expert which took several days, the correct settings were implemented. Also, cards not necessary for the simulation were removed from the code and MULSOFT card which is needed to activate single scattering in thin layers (low density gases) was also included resulting to good results as shown in Figure 7.  AutoFLUKA was able to reproduce the expert’s answers more concisely in a few seconds, while  responses from ChatGPT  on the same questions were general as higlighted in \ref{App_DETEC_spikes}, further demonstrating the potential of RAG in advancing research and development at a fast pace.

\begin{scriptsize}  
\begin{longtable}{|p{13cm}|}
\caption{DETECT Card - Spikes in the Spectrum. The same question from the Forum was asked to AutoFLUKA, which was granted access to the manual, discussions on the topic, and a logbook of knowledge on the topic.} \\ 
\label{Table9_DETECT_spikes}

\textbf{INPUT} \\ \hline

\textbf{User query}: Dear AutoFLUKA, I'm simulating using FLUKA the energy deposition in a low-pressure (low-density) Tissue-equivalent Propane gas. I'm using a 62 MeV proton (point source) and scoring the deposition events via the DETECT card. However, \textbf{the obtained spectrum presents with some spikes on the main peak}. At first, I thought it was a binning problem with the DETECT card, so I changed from the default 1024 to 2000 bins and still got the same results. Then I added the EMFCUT, EMFFIX, and DELTARAY Cards with production/transport threshold energies of 1E-06GeV for electrons and 1E-07GeV for protons. I also added the PART-THR card with a threshold of 1E-14 and still got the spikes. Using your knowledge base with additional documentation from this directory: \texttt{C:/Users/.../FLUKA\_Repository\_user\_forum}, what is the reason (physics) behind these spikes and is there a way to get rid of them? What other FLUKA cards or settings of the above FLUKA input cards can I use? Also, what is the minimum threshold energy for electron production and transport in FLUKA? Please be specific but detailed as well. Where necessary, provide equations and references, including document title(s) and page numbers. \\ \hline

\textbf{OUTPUT} \\ \hline

\textbf{AI Message}: 
The spikes in the energy deposition spectrum you are observing when using the DETECT card in FLUKA for a 62 MeV proton in low-pressure Tissue Equivalent-Propane gas can be attributed to a few factors: \newline
\newline
1. \textbf{Stopping of Low-Energy Electrons}: One common cause of such spikes is the complete stopping of low-energy electrons within the detector. \newline
\texttt{For instance, a peak at about 35 keV/μm can be due to 1 keV electrons stopping entirely inside the detector.} This is a physical phenomenon where low-energy secondary particles deposit all their energy in a very localized region. \newline
\newline
2. \textbf{Proton Stopping in One Step}: Another possible cause is the behavior of protons in the simulation. Protons of different energies might travel through the gas in just one step, and in the simulation, they might be considered as stopped, leading to discrete energy deposition events that appear as spikes. \newline
\newline
3. \textbf{Binning Issues}: Although you have already tried changing the binning from the default 1024 to 2000 bins, binning can still play a role. Ensure that the binning is fine enough to capture the continuous nature of the energy deposition spectrum without introducing artificial spikes. \newline
\newline
4. \textbf{Threshold Settings}: The threshold settings for electron production and transport can also affect the results. Ensure that the thresholds are set appropriately to capture all relevant interactions without missing low-energy events that could contribute to the spectrum. \newline
\newline
\textbf{Recommendations to Address the Spikes:} \newline
1. Refine Binning: Further, refine the binning to ensure it is not too coarse. \newline
2. Adjust Thresholds: Review and adjust the thresholds for electron production and transport to ensure they are set correctly. \newline
3. Check Simulation Parameters: Ensure that all other simulation parameters are set appropriately and that there are no other sources of numerical artifacts. \newline
4. Minimum Threshold Energy for Electron Production and Transport in FLUKA: The minimum threshold energy for electron production and transport in FLUKA is 1 keV. Ensure that this threshold is implemented correctly in your simulation settings. \newline
\newline
By implementing these changes, you should be able to obtain a more realistic energy deposition spectrum and reduce the artificial spikes observed in your results. \\ \hline

\textbf{Follow up}: So, what other Physics Cards can I use to address the above issue and which card(s) might not be so useful in my simulation? You can also use your knowledge together with this documentation: \texttt{C:/Users/.../FLUKA\_Repository\_user\_forum} to answer this follow-up question. \newline
\newline
To address the spikes in the energy deposition spectrum using the DETECT card in FLUKA, you can follow these steps based on the advice provided by Daniele Calolzari: \newline
\newline
\textbf{Useful Physics Cards}: \newline
1. \textbf{EMFCUT Card}: Ensure that you set both the transport and production cuts for electrons to the minimal value. This helps in achieving better precision. \newline
\texttt{EMFCUT   1E-06   1E-7   BLKBODY GLASSTRE} \newline
\texttt{EMFCUT   1E-06   1E-5   BLKHOLE GLASTMAT PROD-CUT} \newline
2. \textbf{DELTARAY Card}: Set the DELTARAY card to the same threshold energy for electron production. This card is crucial as it also sets a shorter step size for protons. \newline
\texttt{DELTARAY  1E-6    BLKHOLE GLASTMAT} \newline
3. \textbf{FLUKAFIX Card}: Optionally, you can manually require shorter step sizes for protons, although this is not generally encouraged as FLUKA is a condensed history Monte Carlo code and may not handle very short step sizes in diluted gases well. \newline
\texttt{FLUKAFIX   0.001   Tissue-E} \newline
4. \textbf{MULSOFT Card}: Although not strictly necessary, it is recommended to switch on single scattering when travelling in thin layers. This can help in achieving a more realistic spectrum. \newline
\texttt{MULSOFT} \newline
\newline
\textbf{Cards That Might Not Be Useful:} \newline
1. \textbf{PART-THR Card}: This card sets the threshold for particle transport, but if you have already set appropriate thresholds using EMFCUT and DELTARAY, this card might not provide additional benefits. \newline
2. \textbf{EMFFIX Card}: This card fixes the energy cut-off for electromagnetic processes, but if you have already set the thresholds using EMFCUT, this card might be redundant. \newline
\newline
By implementing these suggestions, you should be able to obtain a smoother, more realistic energy spectrum. \\ \hline

\textbf{References:}
For detailed information, you can refer to the following documents from your FLUKA Repository:
\begin{itemize}
    \item FLUKA User Manual: This manual provides comprehensive details on the usage of various FLUKA cards, including the DETECT, EMFCUT, EMFFIX, DELTARAY, and PART-THR cards.
    \item FLUKA User Forum: The user forum contains discussions and solutions from the FLUKA user community, which can be very helpful in troubleshooting issues with spikes in the energy spectrum.
\end{itemize}
If you need specific FLUKA-based advice or calculations, please let me know, and I can assist further. \\ \hline

\end{longtable}
\end{scriptsize}

\subsubsection{Uncertainty Propagation from the DETECT Card Spectra}
\label{subsubsection4.2.2}

To compute the radiation quality, the spectra from the DETECT card is converted to the frequency and dose distributions of the lineal energy.  Propagating the uncertainty of the average energy deposited in the detector’s Tissue-equivalent gas sensitive volume to that of the lineal energy and finally the radiation quality is not trivial.  Here, AutoFLUKA demonstrated exceptionalism  in combining the experts’ answers from the User Forum with material from our logbook to generate concise responses to the queries and follow-up questions  in just a few seconds.  Due to the numerous equations retrieved as well as generated by AutoFLUKA, these conversations have been reported in \ref{App_Uncertainty_propagation} and \ref{App_Q_ICRU_uncertainty}. 

ChatGPT’s response to these queries were general even in situations where it had access to the discussions from the FLUKA User Forum.\textit{ Noticeably, ChatGPT provided  a wrong formula for calculating the average lineal energy as highlighted in \ref{App_Uncertainty_propagation}, while AutoFLUKA through this \textbf{fluka\_assistant\_tool} utilized the material it was given as a base to generate an adequate set of formulae for calculating the radiation quality factor}, showcasing the potential of RAG. 

These recommendations, together with others not shown here were adopted to fine-tune our detector codes and to develop post-processing tools for the domain-specific (Microdosimetry) case which will be discussed in the next section. 

\subsection{AutoFLUKA on Microdosimetry }
\label{subsection4.3}

In this case study, we demonstrated AutoFLUKA’s ability to automate the simulation workflow related to the “Design and optimization of an Avalanche-Confinement Tissue Equivalent Proportional Counter (AcTEPC)”. 

The motivation behind this research stems from the fact that the same absorbed dose from different fields of ionizing radiation (IR) can produce different effects on biological targets due to the random nature of radiation interaction with matter. Such interactions are currently considered to be the starting point of radiation-induced damage (Rucinski et al., 2021), which, if left unchecked can become carcinogenic. The same interactions are also exploited to treat cancer and other illnesses in radiation therapy. Therefore, to ensure efficacy while reducing side effects in radiation therapy as well as designing better radiation protection and monitoring systems, radiation detectors capable of precisely measuring such interactions, (i.e. the local energy depositions and ionization clusters in subcellular tissue targets like DNA strands and chromatin fibres) which are important inputs to calculating the \textbf{radiation quality} are needed. Unfortunately, most conventional dosimeters can only measure the absorbed dose, which is a macroscopic averaged quantity that disregards these stochastic effects.  

\begin{figure}[htbp]
    \centering
    \begin{subfigure}[!t]{0.48\linewidth} 
        \centering
        \includegraphics[width=\linewidth]{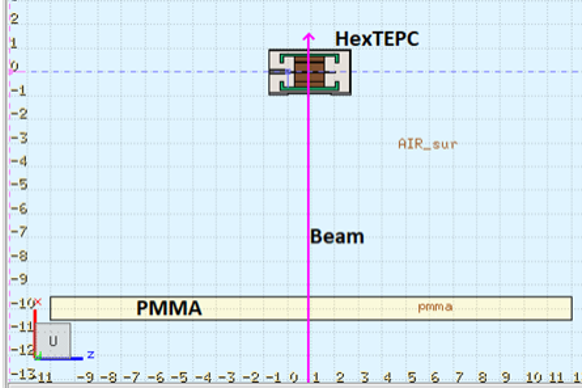}
        \caption{}
        \label{Figure8_a.png}
    \end{subfigure}
    \hfill
    \begin{subfigure}[!t]{0.48\linewidth} 
        \centering
        \includegraphics[width=\linewidth]{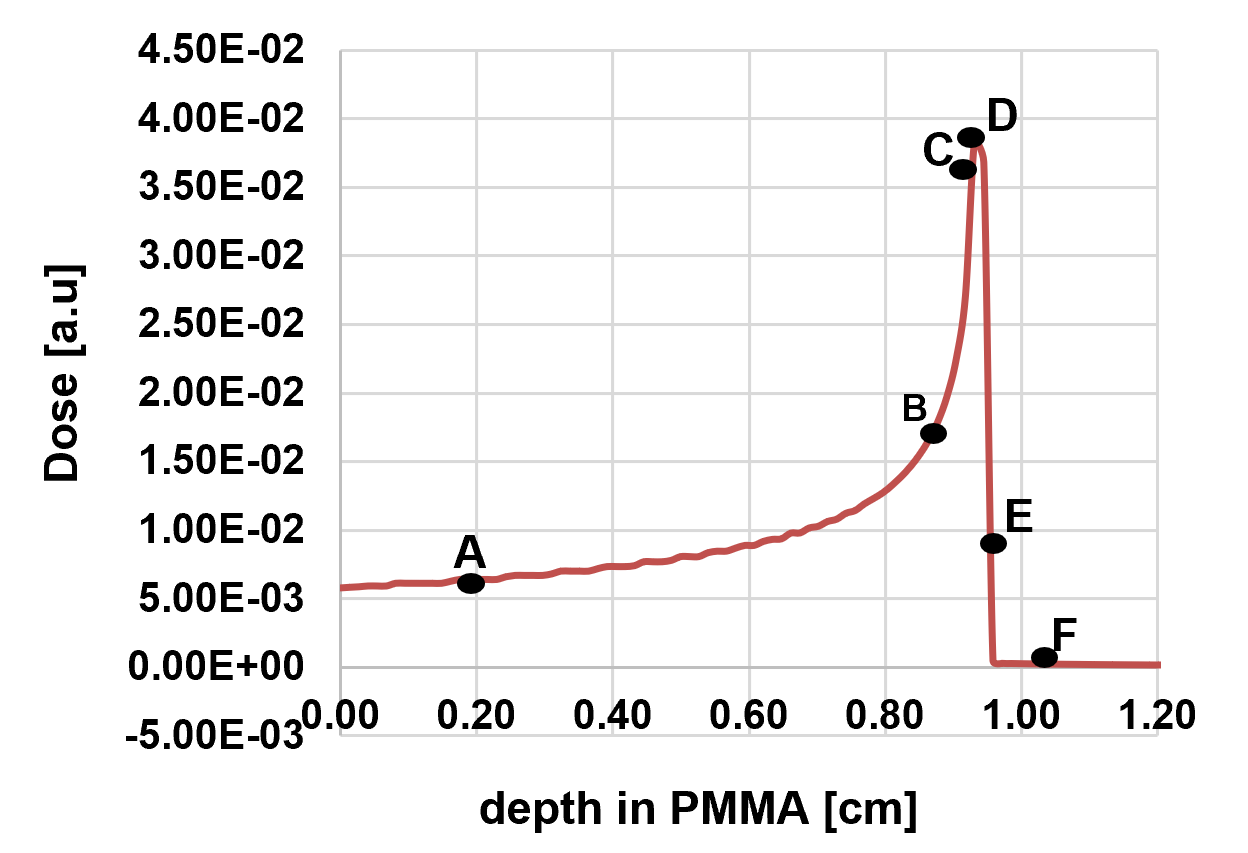}
        \caption{}
        \label{Figure8_b.png}
    \end{subfigure}
    
    \caption{a) - Irradiation scenario in FLUKA for the Design of and optimization of HexTEPC ; (b) Different irradiation positions across the Bragg peak for a 62 MeV/u carbon ion beam passing through Polymethyl methacrylate or PMMA (Ndum et al., 2024).}
    \label{fig8:two-images}
\end{figure}

Through a series of numerical studies, mathematical formulations and MC simulations, we demonstrated that our detector which employs an unconventional (hexagonal) geometry of the gas-sensitive volume (SV), called hexagonal Tissue Equivalent Proportional Counter (HexTEPC)  measure radiation damage in simulated tissue sites ranging from 500 nm down to about 50 nm in diameter. Figure \ref{Figure8_a.png} shows the simulation setup in FLUKA. The brown region is the detector sensitivity volume which simulated an equivalent tissue site. The energy deposited in this region recorded by the DETECT card is converted to the microdosimetric spectra from which the radiation quality is computed. A detailed description can be found here \citep{Ndum2024DesignApplications}.

Optimizing a detector like the HexTEPC involves a time-consuming, error-prone process of running multiple cases with different design parameters like detector wall materials, gas composition and cavity size, etc, in addition to the different settings of the production and transport (physics cards) and scoring cards for energy deposition and other quantities of interest like the source fluence, particle yield, etc). So, automating the workflow saved us a tremendous amount of time, labor and mitigated chances of human errors. 

For brevity, only important parts of the input and output have been shown in Table 10 and  Table 11 respectively. The system message which was hard-coded in to the scheme remained the same as in section 4.1 (Table 6), while the prompt or user query was modified to reflect the current task. As shown, 11 parameters were set by the user; a PMMA thickness of 1.24 cm (6.24 – 5)cm, representing a point beyond F on the Bragg peak from Figure \ref{Figure8_b.png} and a gas density of 3.84615E-06 g/cc representing a simulated tissue diameter (dt) of 50 nm. The DETECT card was configured to record interaction events with energy deposition between 1eV or 1E-09 GeV (E\_min) to 100 keV  or 0.0001 GeV (E\_max). The primary beam’s (carbon ions) energy and position were also set to 62 MeV/u and -10 cm from the detector axis respectively. 

\begin{scriptsize}
\begin{longtable}{|p{13cm}|}
\caption{AutoFLUKA's input (user query). The system message was hard coded into agent using LangChain's ChatPromptTemplate method, while the user query or the prompt was modified to reflect the current task.} \\

\hline
\textbf{INPUT (User Query)} \\ \hline
Please, execute the following workflow in a chronological order from steps 1 through 9. Where needed in the tools, set dt = 50, clf = 2/3 and flag = 0. \newline
Step 1: Read and modify the contents of C:/Users/.../HexTEPC\_Clones\_template.inp with parameters from this csv file: C:/Users/.../hextepc\_params.csv. \newline
Step 2: Create 10 fluka input copies with prefix "hex\_tepc" in the directory C:/Users/.../Results. \newline
Step 3: When all the input files have been created, execute FLUKA, AND decrypt all data from output files ending in [\_fort.xx], where [xx] is any number between 17 through 99. \newline
Step 4: Next, process the resulting data from output files ending in [\_sum.lis] and [\_tab.lis] and save results as [fluka\_data.json]. Extract the [Average Uncertainty] from the [output\_fort\_17\_tab.lis] tab\_section AND the [Total Primaries Run] or nps from the [output\_fort\_50\_sum.lis] sum\_section of the first detector from this [fluka\_data.json] data. \newline
Step 5: \textbf{REMEMBER!} If this Average Uncertainty is less than 2\%, then GO TO Step 8. Else, GO TO Step 6. \newline
Step 6: Given that the Average Uncertainty is inversely proportional to the square root of the nps, what value of nps (rounded to the next 100000th) is needed to get an uncertainty less than 2\%? \newline
Step 7: Next, update the csv file C:/Users/.../hextepc\_params.csv with this new nps, then REPEAT steps 1 to 4, SKIP STEP 5 and CONTINUE from step 8. \newline
Step 8: Next, load the 'fluka\_data.json' file, SELECT ONLY the [output\_fort\_17\_tab.lis] data and perform a logarithmic rebinning, saving the log-rebinned data as 'tepc\_log\_data.json'. \newline
Step 9: Load the 'tepc\_log\_data.json', Compute and plot the microdosimetric spectra data. When done, GO TO FINISH. \\ \hline

\end{longtable}  
\end{scriptsize}

\begin{figure}
    \centering
    \includegraphics[width=1\linewidth]{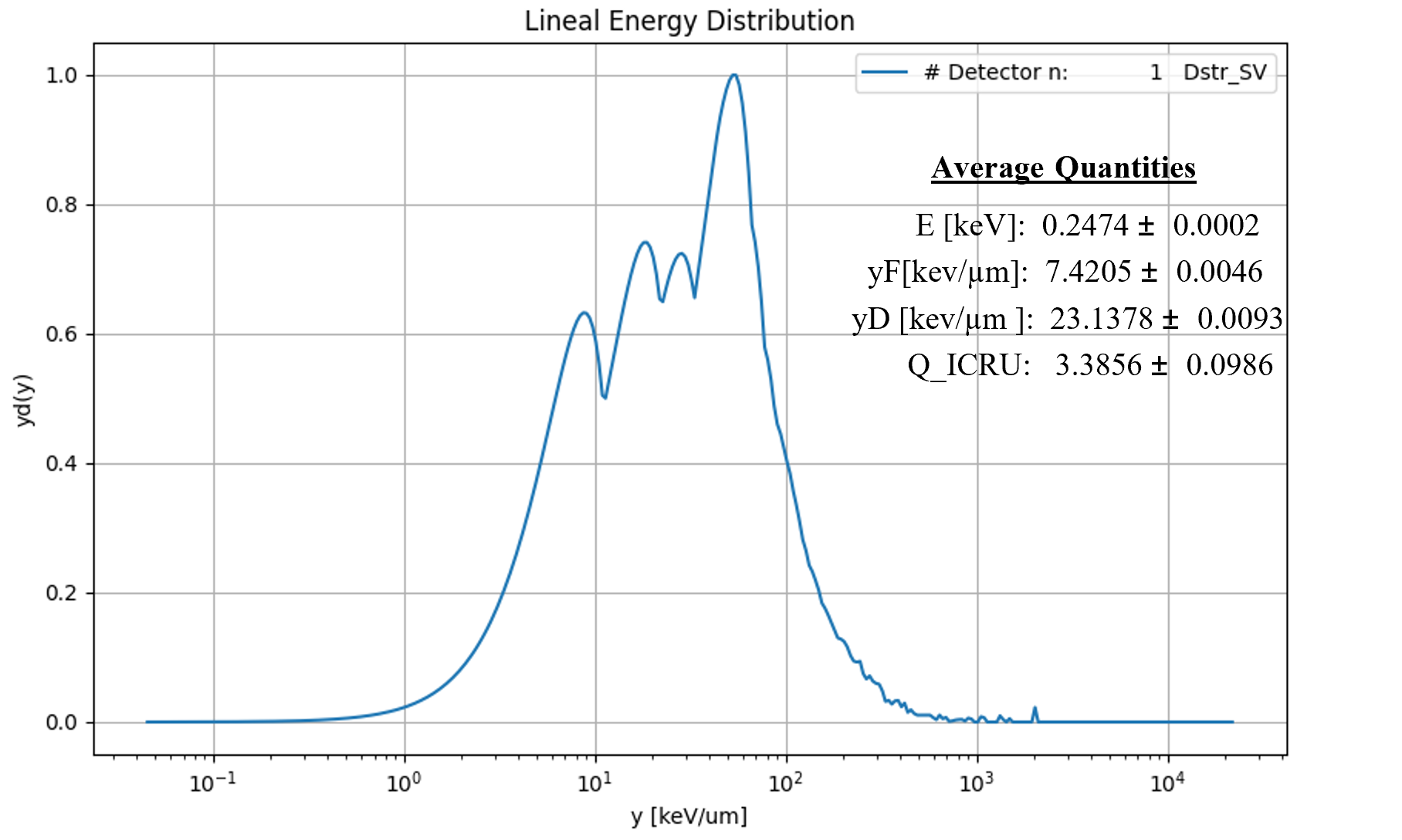}
    \caption{Dose distribution of the lineal energy generated by the  microdosimetric\_spectra\_tool. E represents the Average energy deposited in the Detector’s sensitive volume which simulates the Tissue site,  yF represents the frequency-mean of the lineal energy while  yD represents the dose-mean of the lineal energy.}
    \label{fig:dose-distr}
\end{figure}

The comments generated by AutoFLUKA during this execution process like the one shown for recalculating the NPS suggest that the agent clearly understood the task. The corresponding microdosimetric spectrum (dose distribution of the lineal energy) generated by the microdosimetric spectra tool is shown in Figure \ref{fig:dose-distr}. For the quantities shown, E represents the average energy deposited in the Detector’s sensitive volume, which simulates the tissue site, while yF and yD respectively represent the frequency-mean and the dose-mean of the lineal energy (y). The Quality factor calculated following recommendations from the International Commission on radiological Measurements Report 98 \citep{Braby2023ICRUMicrodosimetry} is also shown. 

\begin{scriptsize}
\begin{longtable}{|p{2.1cm}|p{11.2cm}|}
\caption{AutoFLUKA’s output. The generated AI messages show that the task was clearly understood.} \\
\hline

\multicolumn{2}{|c|}{\textbf{OUTPUT}} \\ \hline

\textbf{CSV file reading} & 
\ttfamily 
\textbf{TOOL: csv\_file\_reader\_tool $\rightarrow$ call\_pUWAwG65vJL97tSXTxiic6aR}
\begin{multicols}{2}
beam\_p\_gev: "-0.062" \newline
beam\_pos\_x: "0" \newline
prima\_thickness: "6.24" \newline
clf: "0.667" \newline
\columnbreak 
density: "3.84615e-06" \newline
E\_min: "1e-09" \newline
E\_max: "10.000" \newline
Nbins: "2000" \newline
seed: "10" \newline
nps: "1000"
\end{multicols}
\normalfont \\ \hline

\textbf{Code execution} & \textbf{TOOL: nps\_and\_uncertainty\_tool $\rightarrow$ call\_8WoL8NstX5INQweTliYjzLg} \newline
Execution Directory: /mnt/c/.../HexTEPC/Results \newline
How Many Input files? 10 \newline
Now Running:  \newline
AutoFLUKA\_job1.sh for : hex\_tepc\_01.inp \newline
AutoFLUKA\_job2.sh for : hex\_tepc\_02.inp \newline
... \newline
AutoFLUKA\_job9.sh for : hex\_tepc\_09.inp \newline
AutoFLUKA\_job10.sh for : hex\_tepc\_10.inp \newline
All Jobs are running... \newline
Simulations successfully executed at: 2024-09-15 15:18:39 \newline
Simulation Time: 00:29:53.013917 \\ \hline

\textbf{Uncertainty and NPS extraction} & \textbf{TOOL: nps\_and\_uncertainty\_tool $\rightarrow$ call\_8WoL8NstX5INQweTliYjzLg} \newline
1. Average Uncertainty: 10.08974358974364 \newline
2. Total Primaries Run: 10000 \\ \hline

\textbf{Recalculation of NPS and update of CSV parameters file} & AI \newline
The required nps to achieve an uncertainty less than 10\% is 400,000. \newline
Step 7: Update the CSV file with the new nps and repeat steps 1 to 4, skipping step 5. \newline
We'll update the CSV file with the new nps value and repeat the necessary steps. \newline
Update the CSV file \newline
Let's update the CSV file with the new nps value. \\ \hline

\textbf{Re-run the Simulations} & \textbf{TOOL: fluka\_executer\_tool $\rightarrow$ call\_rzkl21xyBcsUNakOLRCGrYNtTE} \newline
Execution Directory: /mnt/c/.../HexTEPC/Results \newline
How Many Input files? 10 \newline
Now Running: \newline
AutoFLUKA\_job1.sh for : hex\_tepc\_01.inp \newline
AutoFLUKA\_job2.sh for : hex\_tepc\_02.inp \newline
... \newline
AutoFLUKA\_job9.sh for : hex\_tepc\_09.inp \newline
AutoFLUKA\_job10.sh for : hex\_tepc\_10.inp \newline
All Jobs are running... \newline
Simulations successfully executed at: \newline
Simulation Time: ... \\ \hline

\textbf{Log rebinning on output data} & \textbf{TOOL: lin\_to\_log\_rebinning\_tool $\rightarrow$ call\_AK2SpRces4LiKnNUOCgTjlJs7} \newline
Rebinned data saved as \newline
C:/Users/.../Results/tepc\_data.json \\ \hline

\textbf{Radiation Quality Computation} & \textbf{TOOL: microdosimetric\_spectra\_tool $\rightarrow$ call\_gXcPMTDo5LhWbccpBy4W} \newline
file\_path: C:/Users/.../HexTEPC/Results \newline
json\_filename: tepc\_data.json \newline
dt: 300 \newline
clf: 0.6666666666666666 \\ \hline

\textbf{Summary of steps} & AI \newline
\textbf{Workflow Complete} \newline
The microdosimetric spectra data has been computed and plotted successfully. The workflow has been completed as expected. \newline
If you need any further assistance or analysis, please let me know! \\ \hline

\end{longtable}
\end{scriptsize}

As mentioned earlier, the stepwise approach adopted for the user prompts enabled us to tweak the agents to commence anywhere in the workflow. For instance, if we wanted to analyze results from previous FLUKA simulations, we could ask AutoFLUKA to search and decrypt the binary files from the specified directory, create the necessary Json file(s) and finally plot the data for visualization as depicted in Appendix --

\section{CONCLUSION}
\label{sec5}

In this study, we successfully demonstrated the application of AI agents to automate Monte Carlo simulation workflows in FLUKA. Traditionally, the complexity of Monte Carlo simulations involves numerous parameters, a steep learning curve, and extensive manual intervention. These factors, in turn, lead to cumbersome processes that are time-consuming, prone to human error, and often require a significant investment of both time and computational resources. AutoFLUKA addresses these challenges by streamlining and automating key aspects of the workflow, thereby mitigating common issues such as human error, loss of execution time, and the inefficient use of computational resources.

Through a carefully crafted sequence of prompts, AutoFLUKA enables users to modify FLUKA input files, execute simulations, and effectively manage post-processing tasks, all with a single command. This not only saves time but also ensures the precision of the simulation by reducing the possibility for manual errors. Furthermore, the incorporation of tools to decrypt and store output data in structured JSON format facilitates easier downstream analysis and integration, enhancing both the usability and utility of the results.

One of the significant achievements of AutoFLUKA is its capability to automate the generation of multiple statistically independent input files. This feature is particularly valuable for scenarios where reducing simulation time is crucial while maintaining good statistical quality. AutoFLUKA also provides users with flexibility in modifying critical parameters, ensuring that specific configurations can be tested efficiently without manual file duplication and editing.

Moreover, the introduction of the FLUKA assistant tool based on RAG shows promise in transforming user experience. By augmenting the AI model with domain-specific data, such as the FLUKA user manual and expert discussions, AutoFLUKA can assist users by addressing straightforward queries related to simulation setup, output analysis, and error troubleshooting. This feature can significantly reduce the dependency on the FLUKA user forum, potentially saving both time for users and workload for FLUKA experts.

Future work will focus on refining the capabilities of AutoFLUKA by extending the RAG tool to handle a broader range of data, including figures and graphs, which are often crucial for technical documentation. Further improvements in the multi-agent framework are also envisioned, potentially automating more complex human interventions such as ensuring adherence to input file formatting requirements. With these advancements, AutoFLUKA could serve as a versatile tool for any simulation software that relies on text-based input files, opening new avenues for research and development in nuclear science and engineering. This approach can be applied to any simulation software with a text-based input file system, such as the Monte Carlo N-Particle (MCNP) code, Particle and Heavy Ion Transport Code System (PHITS), etc. 


\appendix
\section{DETECT Card - Spikes in the Spectrum}
\label{App_DETEC_spikes}

The discussion is presented in section \ref{subsec4.2} and Table \ref{Table9_DETECT_spikes}.

\noindent
\textbf{Comparison with ChatGPT}

The discussion can be viewed at this \href{https://chatgpt.com/share/e/66eba6ab-1b4c-8006-9b1b-f1b9f9607bfb}{link}

\noindent
\textbf{Comments}

Even with access to the FLUKA User Forum discussions (not shown here), ChatGPT gave a rather generic response to the query with no specific code examples (Card configurations).

\section{Uncertainty Propagation from the DETECT Card’s Spectra}
\label{App_Uncertainty_propagation}

\noindent
\textbf{User query:}\\
Hi AutoFLUKA, I am designing a Tissue Equivalent Proportional Counter (TEPC) using the Monte Carlo Particle Transport code FLUKA. I score the energy deposited in the detector's gas Sensitive Volume (SV) using the DTECT card, which has columns of data: lower energy bin (LB), upper energy bin (UB), Counts (C), and Uncertainty (U). According to your knowledge and documentation, how can I propagate the average uncertainty from this DETECT Card to the lineal energy? Please provide a detailed summary, including equations and references, if necessary.\\

\noindent
\textbf{AutoFLUKA's Response}\\
To propagate the average uncertainty from the DETECT card to the lineal energy, follow this systematic approach:

\noindent
\textbf{Step-by-Step Guide}\\
\noindent
1. \textbf{Calculate the Average Energy and Uncertainty from DETECT Card:}
   - Use the recorded spectra to calculate the average energy.
   - Ensure an appropriate \(E_{\text{minimum}}\) is set.
   - Calculate the average energy (\( \bar{E} \)) and its uncertainty \( \sigma_E \).\\
\noindent
2. \textbf{Convert to Lineal Energy:}
   - Lineal energy is the energy deposited per unit length.
   - Calculate lineal energy (\( \bar{y} \)) as: 
   \[
   \bar{y} = \frac{\bar{E}}{l}
   \]
\noindent
3. \textbf{Propagate the Uncertainty:}
   - The uncertainty in lineal energy \( \sigma_y \) is propagated as:
   \[
   \sigma_y = \frac{\sigma_E}{\hat{l}}
   \]
   where, \(\hat{l}\) is the mean chord length. \\
\noindent
4. \textbf{Summarize the Results:}
   - Report the average lineal energy (\( \bar{y} \)) and its propagated uncertainty \( \sigma_y \).\\

\noindent
\textbf{Follow-up Question:}\\
What formula is used to calculate the average energy above?\\

\noindent
\textbf{AutoFLUKA's Response}\\
The formula for average energy \( \langle E \rangle \) is:

\[
\langle E \rangle = \frac{\sum_{i=1}^{N} E_i \cdot C_i}{\sum_{i=1}^{N} C_i}
\]

Where:
- \( E_i \) is the midpoint of the \( i \)-th energy bin.
- \( C_i \) is the count in the \( i \)-th energy bin.
- \( N \) is the total number of bins.

\noindent
\textbf{Steps to Calculate Average Energy:}\\
1. \textbf{Determine the Midpoint of Each Energy Bin:}
   \[
   E_i = \frac{LB_i + UB_i}{2}
   \]
   
\noindent  
2. \textbf{Calculate the Weighted Sum of Midpoints:}
   \[
   \sum_{i=1}^{N} E_i \cdot C_i
   \]

\noindent
3. \textbf{Calculate Total Counts:}
   \[
   \sum_{i=1}^{N} C_i
   \]

\noindent
4. \textbf{Compute the Average Energy:}
   \[
   \langle E \rangle = \frac{\sum_{i=1}^{N} E_i \cdot C_i}{\sum_{i=1}^{N} C_i}
   \]

\noindent
\textbf{Example Calculation:}

\begin{center}
\begin{tabular}{|c|c|c|}
\hline
\textbf{Lower Bound (LB)} & \textbf{Upper Bound (UB)} & \textbf{Counts (C)} \\
\hline
0.0 & 0.1 & 10 \\
0.1 & 0.2 & 20 \\
\hline
\end{tabular}
\end{center}

\noindent
1. \textbf{Calculate Midpoints:}
\[
E_1 = \frac{0.0 + 0.1}{2} = 0.05
\]
\[
\noindent
E_2 = \frac{0.1 + 0.2}{2} = 0.15
\]

\noindent
2. \textbf{Calculate the Weighted Sum of Midpoints:}
\[
\sum_{i=1}^{N} E_i \cdot C_i = (0.05 \cdot 10) + (0.15 \cdot 20) = 0.5 + 3.0 = 3.5
\]

\noindent
3. \textbf{Calculate Total Counts:}
\[
\sum_{i=1}^{N} C_i = 10 + 20 = 30
\]

\noindent
4. \textbf{Compute the Average Energy:}
\[
\langle E \rangle = \frac{3.5}{30} = 0.1167
\]

Thus, the average energy deposited is 0.1167 units.

\noindent
\textbf{Comparison with ChatGPT (Zero-shot):}\\
The discussion can be viewed at this \href{https://chatgpt.com/share/e/66eda788-b6f8-8006-b03d-a8e62d8267f4}{link}.

\noindent
\textbf{Comments:}\\
ChatGPT missed the calculation of the average lineal energy by a factor of \(1/\hat{l}\) (reciprocal of the mean chord length).

\section{Uncertainty of the Radiation Quality Factor: $Q_{\text{ICRU}}$}
\label{App_Q_ICRU_uncertainty}

\noindent
\textbf{User query:}

Please, provide formulae for the frequency and dose distributions of the lineal energy and the ICRU Radiation Quality factor (\(Q_{\text{ICRU}}\)) and their respective means. Please pay attention to the exponential form of \(Q_{\text{ICRU}}\) given in the documents: C:/Users/.../PDFs\_Temporal.\\

\noindent
\textbf{AutoFLUKA's Response} 

Here are the formulae for the frequency and dose distributions of the lineal energy, the ICRU Radiation Quality factor (\(Q_{\text{ICRU}}\)), and their respective means:

\noindent
\textbf{Frequency Distribution $f(y)$ and Dose Distribution $d(y)$}

\noindent
1. \textbf{Frequency Distribution $f(y)$:}
\[
f(y) = \frac{C_i}{\Delta y_i \sum_i C_i}
\]
where \(C_i\) is the count in the \(i\)-th bin, and \(\Delta y_i\) is the width of the \(i\)-th bin.

\noindent
2. \textbf{Dose Distribution $d(y)$:}
\[
d(y) = \frac{y f(y)}{y_F}
\]
where \(y_F\) is the average lineal energy, given by:
\[
y_F = \frac{\int y^2 f(y) dy}{\int y f(y) dy}
\]

\noindent
\textbf{Mean Lineal Energies}\\
\noindent
1. \textbf{Average Lineal Energy}:
\[
\bar{y} = \int y f(y) dy
\]

\noindent
2. \textbf{Dose-Mean Lineal Energy}:
\[
\bar{y}_D = \frac{\int y^2 f(y) dy}{\int y f(y) dy}
\]

\noindent
\textbf{ICRU Radiation Quality Factor (Q\textsubscript{ICRU})}

\noindent
1. \textbf{ICRU Radiation Quality Factor ($Q_{\text{ICRU}}$)}:
    \[
    Q_{\text{ICRU}}(y) = 5.60 \times 10^{-5} y^2 \left( 1 - e^{-0.5x \times 10^{-4}} - 2 \times 10^{-6}y \right)
    \]

\noindent
2. \textbf{Mean Quality Factor ($Q_{\text{ICRU, avg}}$)}:
\[
Q_{\text{ICRU, avg}} = \frac{\int Q_{\text{ICRU}}(y) d(y)}{\int d(y)}
\]

These formulae provide a comprehensive framework for calculating the frequency and dose distributions of lineal energy, as well as the ICRU Radiation Quality factor and their respective means.\\

\noindent
\textbf{Follow-up Question:} \\
Now, assuming that the average quality is calculated as a summation rather than an integral like so:
\[
Q_{\text{ICRU, avg}} = \frac{\text{np.sum}(Q_i \ast d_i)}{\text{np.sum}(d_i)}
\]
and your knowledge about uncertainty propagation, \textbf{GIVEN IN THE DOCUMENTS:} C:/Users/.../PDFs\_Temporal, please derive the expression for the uncertainty of the average quality factor. 

Please note the following: The \textbf{ONLY} uncertainties present are that from the raw counts \(C_i\) and the frequency-mean lineal energy \(y_F\). To do this, first start from the known uncertainties (\(\sigma_{C_i}\) and \(\sigma_{y_F}\)) then:
1) Calculate the uncertainty of the frequency distribution (\(\sigma_{f_i}\)).
2) Propagate it to the uncertainty of the dose distribution (\(\sigma_{d_i}\)). The lineal energy (\(y\)) has \textbf{NO} uncertainty.
3) Finally, propagate this to the average quality factor uncertainty.

When done, please make a summary and provide a python script to automatically calculate this, given that \(\sigma_{C_i}\) and \(\sigma_{y_F}\) are given.\\

\noindent
\textbf{AutoFLUKA's Response}

To derive the expression for the uncertainty of the average quality factor, we will follow the steps outlined:

\noindent
\textit{Step 1: Calculate the Uncertainty of the Frequency Distribution ($\sigma_{f_i}$)}
The frequency distribution \(f(y)\) is given by:
\[
f(y) = \frac{C_i}{\Delta y_i \sum_i C_i}
\]
The uncertainty in \(f(y)\), denoted as \(\sigma_{f_i}\), can be calculated using the propagation of uncertainty formula:
\[
\sigma_{f_i} = f(y) \sqrt{\left(\frac{\sigma_{C_i}}{C_i}\right)^2 + \left(\frac{\sigma_{\Sigma_i C_i}}{\sum_i C_i}\right)^2}
\]
Since \(\sigma_{\Sigma_i C_i}\) is the combined uncertainty of all counts:
\[
\sigma_{\Sigma_i C_i} = \sqrt{\sum_i \sigma_{C_i}^2}
\]

\noindent
\textit{Step 2: Propagate to the Uncertainty of the Dose Distribution ($\sigma_{d_i}$)}
The dose distribution \(d(y)\) is given by:
\[
d(y) = \frac{y f(y)}{y_F}
\]
The uncertainty in \(d(y)\), denoted as \(\sigma_{d_i}\), can be calculated using the propagation of uncertainty formula:
\[
\sigma_{d_i} = d(y) \sqrt{\left(\frac{\sigma_{f_i}}{f_i}\right)^2 + \left(\frac{\sigma_{y_F}}{y_F}\right)^2}
\]

\noindent
\textit{Step 3: Propagate to the Average Quality Factor Uncertainty}
The average quality factor \(Q_{\text{ICRU, avg}}\) is given by:
\[
Q_{\text{ICRU, avg}} = \frac{\sum_i Q_i d_i}{\sum_i d_i}
\]
The uncertainty in \(Q_{\text{ICRU, avg}}\), denoted as \(\sigma_{Q_{\text{ICRU, avg}}}\), can be calculated using the propagation of uncertainty formula:
\[
\sigma_{Q_{\text{ICRU, avg}}} = \sqrt{\sum_i \left(\frac{\partial Q_{\text{ICRU, avg}}}{\partial d_i} \sigma_{d_i}\right)^2}
\]
The partial derivative of \(Q_{\text{ICRU, avg}}\) with respect to \(d_i\) is:
\[
\frac{\partial Q_{\text{ICRU, avg}}}{\partial d_i} = \frac{Q_i \sum_j d_j - \sum_j Q_j d_j}{(\sum_j d_j)^2}
\]
Substituting this into the uncertainty propagation formula, we get:
\[
\sigma_{Q_{\text{ICRU, avg}}} = \sqrt{\sum_i \left(\frac{Q_i \sum_j d_j - \sum_j Q_j d_j}{(\sum_j d_j)^2} \sigma_{d_i}\right)^2}
\]

\noindent
\textbf{Summary}

1. Calculate \(\sigma_{f_i}\) using the uncertainties in the counts \(\sigma_{C_i}\).
2. Propagate \(\sigma_{f_i}\) to \(\sigma_{d_i}\) using the uncertainty in the frequency-mean lineal energy \(y_F\).
3. Calculate \(\sigma_{Q_{\text{ICRU, avg}}}\) using the propagated uncertainties \(\sigma_{d_i}\).\\

\noindent
\textbf{Python Script}

Here is a Python script to automatically calculate the uncertainty of the average quality factor given \(\sigma_{C_i}\) and \(\sigma_{y_F}\):

\begin{scriptsize}   
\begin{verbatim}
import numpy as np

def calculate_uncertainty(C_i, Q_i, y_F, sigma_C_i, sigma_y_F):
    # Calculate frequency distribution f_i
    sum_C_i = np.sum(C_i)
    f_i = C_i / sum_C_i
    
    # Calculate uncertainty in frequency distribution sigma_f_i
    sigma_sum_C_i = np.sqrt(np.sum(sigma_C_i**2))
    sigma_f_i = f_i * np.sqrt((sigma_C_i / C_i)**2 + (sigma_sum_C_i / sum_C_i)**2)
    
    # Calculate dose distribution d_i
    d_i = y_F * f_i / y_F
    
    # Calculate uncertainty in dose distribution sigma_d_i
    sigma_d_i = d_i * np.sqrt((sigma_f_i / f_i)**2 + (sigma_y_F / y_F)**2)
    
    # Calculate average quality factor Q_ICRU_avg
    Q_ICRU_avg = np.sum(Q_i * d_i) / np.sum(d_i)
    
    # Calculate uncertainty in average quality factor sigma_Q_ICRU_avg
    sum_d_i = np.sum(d_i)
    partial_derivative = (Q_i * sum_d_i - np.sum(Q_i * d_i)) / (sum_d_i**2)
    sigma_Q_ICRU_avg = np.sqrt(np.sum((partial_derivative * sigma_d_i)**2))
    
    return Q_ICRU_avg, sigma_Q_ICRU_avg
\end{verbatim}
\end{scriptsize}
\noindent
\textbf{Comments}

This Python function was modified and incorporated into the tools to compute $Q_{\text{ICRU}}$ and its associated uncertainty.

\section{Multi-Agent Workflow}
\label{AppD}

Here, we show similar results gotten using the multi-agent approach. 

\begin{scriptsize}
\begin{longtable}{|p{1.3cm}|p{11.5cm}|}
\caption{AutoFLUKA's task assignment with multiple workers to process the given query.}\\
\hline
\multicolumn{2}{|c|}{\textbf{INPUT}} \\ \hline

\textbf{System Message} & 
You are a supervisor tasked with managing tasks between the following workers: PythonREPL, text\_file\_reader, csv\_file\_reader, fluka\_input, fluka\_executer, fluka\_data\_decrypter, fluka\_data\_to\_json, nps\_and\_uncertainty, fluka\_data\_plotter, weight\_data\_with\_gas\_gains, lin\_to\_log\_rebinning, microdosimetric\_spectra, fluka\_rag\_assistant.

Given the following user request, respond with the worker to act next. Each worker will perform a task and respond with their results and status. When finished, respond with FINISH. \newline
Please, execute the following workflow in a chronological order from steps 1 through 7. You Have ALL the tools you need to manage tasks between your workers and the parameters they will need. When you encounter an error, take a deep breath. \\ \hline

\textbf{Human (user query)} & 
Please, execute the following workflow in a chronological order from steps 1 through 9:

\textbf{Step 1:} Read the fluka input file located at this directory: \newline
C:/Users/.../example\_template.inp and modify its contents with parameters from the csv file which you will also read from this directory: C:/Users/.../parameters.csv. \newline
\textbf{Step 2:} Next, create 5 fluka input copies with prefix 'example' in the directory: C:/Users/.../Results \newline
\textbf{Step 3:} When all the input files have been created, execute fluka, AND Decrypt all data from output files ending in [\_fort.xx], where [xx] is any number between 17 through 99. \newline
\textbf{Step 4:} Next, process the resulting data from output files ending in [\_sum.lis] and [\_tab.lis] and save results as [fluka\_data.json]. Extract the \textbf{[Average Uncertainty]} from the \textbf{output\_fort\_46\_tab.lis} tab\_section AND the \textbf{[Total Primaries Run]} or nps from the \textbf{output\_fort\_46\_sum.lis} sum\_section of the first detector from this [fluka\_data.json] data. \newline
\textbf{Step 5:} REMEMBER if this Average Uncertainty is less than 10\%, THEN GO TO Step 8. Else, GO TO Step 6. \newline
\textbf{Step 6:} Given that the Average Uncertainty is inversely proportional to the square root of the nps, what value of nps (rounded to the next 100000th) is needed to get an uncertainty less than 10\%? \newline
\textbf{Step 7:} Next, UPDATE the csv file C:/Users/.../parameters.csv with this new nps, then REPEAT steps 1 to 4, SKIP STEP 5 and CONTINUE from step 8. \newline
\textbf{Step 8:} Next, load the \textbf{fluka\_data.json} file and perform a logarithmic rebinning, saving the log-rebinned data as \textbf{tepc\_log\_data.json}. \newline
\textbf{Step 9:} Load the \textbf{tepc\_log\_data.json}, Compute and plot the microdosimetric spectra data. When done with the workflow, GO to \textbf{FINISH}.\\ \hline

\end{longtable}
\end{scriptsize}

\textbf{Workflow Output }

\begin{scriptsize}
\begin{longtable}{|p{2.5cm}|p{10cm}|}
\caption{AutoFLUKA’s output. The generated AI messages show that the task was clearly understood.} \\
\hline

\multicolumn{2}{|c|}{\textbf{OUTPUT}} \\ \hline

\textbf{Message from supervisor:} & \texttt{‘supervisor’: \{ ‘next’: ‘text\_file\_reader’ \}} \\ \hline

\textbf{Step 1 and Step 2 Completed} &
Now Creating Input Files...\newline
The FLUKA input file has been read and modified with parameters from the CSV file. Five copies of the modified FLUKA input file have been created with the prefix 'example' in the specified directory. \newline
\textbf{Next Steps:}\newline
Since I cannot execute FLUKA or decrypt files directly, you will need to perform the following steps manually or with another system:\newline
\\ \hline

\textbf{Step 3: Execute FLUKA and Decrypt Output Files} &
1. Execute FLUKA with the generated input files. \newline
2. Decrypt all data from output files ending in \_fort.xx, where xx is any number between 17 and 99. \\ \hline

\textbf{Step 4: Process Resulting Data} &
1. Process the resulting data from output files ending in \_sum.lis and \_tab.lis. \newline
2. Save the results as fluka\_data.json. \newline
3. Extract the Average Uncertainty from the output\_fort\_46\_tab.lis tab section. \newline
4. Extract the Total Primaries Run (nps) from the output\_fort\_46\_sum.lis sum section of the first detector. \\ \hline

\textbf{Step 5: Check Average Uncertainty} &
1. If the Average Uncertainty is less than 10\%, proceed to Step 8. \newline
2. Otherwise, proceed to Step 6. \\ \hline

\textbf{Step 6: Calculate New nps} &
1. Calculate the new nps needed to achieve an uncertainty less than 10\%. \newline
2. Update the parameters in the CSV file with the new nps. \\ \hline

\textbf{Step 7: Repeat Steps 1 to 4} &
1. Repeat Steps 1 to 4 with the new parameters. \newline
2. Skip Steps 5, 6, and 7, and continue from Step 8. \\ \hline

\textbf{Step 8: Plot Data} &
1. Plot the data from the fluka\_data.json file with the following specifications: \newline
   \texttt{o plot\_error\_bars=False} \newline
   \texttt{o plot\_blocks=False} \newline
   \texttt{o log\_scale=False} \newline
   \texttt{o semilogx=True} \newline
   \texttt{o semilogy=False} \newline
2. Save the results as JPEG images in the same directory. \\ \hline

\textbf{AI Message from next worker:} &
\texttt{‘supervisor’: \{ ‘next’: ‘fluka\_executer’ \}} \newline
\textbf{Executing FLUKA at:} \newline
2024-09-28 21:10:36 \newline
Execution Directory: /mnt/c/Users/.../Results\_MA \\ \hline

\textbf{FLUKA Execution Output} &
How Many Input files? 5 \newline
Now Running: \newline
\texttt{AutoFLUKA\_job1.sh for : example\_01.inp} \newline
\texttt{AutoFLUKA\_job2.sh for : example\_02.inp} \newline
\texttt{AutoFLUKA\_job3.sh for : example\_03.inp} \newline
\texttt{AutoFLUKA\_job4.sh for : example\_04.inp} \newline
\texttt{AutoFLUKA\_job5.sh for : example\_05.inp} \newline
All Jobs are running... \newline
Simulations successfully executed at: \newline
2024-09-28 21:11:35 \newline
\textbf{Simulation Time: 00:00:59.503181} \\ \hline

\textbf{Binary to human-readable} & 
Step 3 Completed: \newline
The FLUKA simulations have been successfully executed. The FLUKA output files have been decrypted successfully. \newline
Now, I will process the resulting data from the output files ending in \_sum.lis and \_tab.lis and save the results as fluka\_data.json. I will extract the Average Uncertainty from the output\_fort\_46\_tab.lis tab section and the Total Primaries Run (nps) from the output\_fort\_46\_sum.lis sum section of the first detector. Let’s proceed with processing the data.\\ \hline

\textbf{Organize data in JSON format} & 
\texttt{\{'supervisor': \{'next': 'fluka\_data\_to\_json'\}\}} \newline
Data has been written to fluka\_data.json.\\ \hline

\textbf{Check Statistical accuracy} & 
Step 5: Check Average Uncertainty \newline
Let’s check the Average Uncertainty from the fluka\_data.json file. The Average Uncertainty is 12.5\%, which is greater than 10\%. Therefore, we need to proceed to Step 6. \newline
Step 6: Calculate New nps \newline
Since the Average Uncertainty is inversely proportional to the square root of nps, we need to calculate the new nps to achieve an uncertainty less than 10\%. The formula to calculate the new nps is: \newline
\[
\text{{nps}} = \left(\frac{{\text{{current uncertainty}}}}{{\text{{desired uncertainty}}}}\right)^2 \times \text{{current nps}}
\]
Let’s calculate the new nps: \newline
\[
\text{{nps}} = \left(\frac{{12.5}}{{10}}\right)^2 \times 1000000
\]
\[
\text{{nps}} = 1.5625 \times 1000000 = 1562500
\]
Rounding to the nearest 100000th: \newline
\[
\text{{new nps}} = 1600000
\]\\ \hline

\textbf{Repeat the simulations} & 
\texttt{\{'supervisor': \{'next': 'csv\_file\_reader'\}\}} \newline
\textbf{Now Creating Input Files...} \newline
Step 1 and Step 2 Repeated: \newline
The FLUKA input file has been read and modified with the updated parameters from the CSV file, including the new nps value of 1600000. Five copies of the modified FLUKA input files have been created with the prefix ‘example’ in the specified directory. \newline
Next Steps: \newline
Since I cannot execute FLUKA or decrypt files directly, you will need to perform the following steps manually or with another system: \newline
Step 3: Execute FLUKA and Decrypt Output Files. \newline
Step 4: Process Resulting Data.\newline

\textbf{...continuous} \newline

\textbf{Once you have the }fluka\_data.json\textbf{ file, I can assist you with plotting the data. Please provide the }fluka\_data.json\textbf{ file when you are ready to proceed with Step 8.} \newline

\\ \hline

\textbf{Repeat the simulations} & 
\ttfamily {'supervisor': {'next': 'fluka\_executer'}} \newline
\textbf{Starting FLUKA at:} \newline
2024-09-28 21:10:36 \newline
Execution Directory: /mnt/c/Users/.../Results\_MA \newline
\textbf{How Many Input files? 5} \newline
Now Running: \newline
AutoFLUKA\_job1.sh for : example\_01.inp \newline
AutoFLUKA\_job2.sh for : example\_02.inp \newline
AutoFLUKA\_job3.sh for : example\_03.inp \newline
AutoFLUKA\_job4.sh for : example\_04.inp \newline
AutoFLUKA\_job5.sh for : example\_05.inp \newline
All Jobs are running... \newline
Simulations successfully executed at: \newline
2024-09-28 21:11:35 \newline
\textbf{Simulation Time: 00:01:03.237136 }\\ \hline

\textbf{Step 4 Completed} & 
The resulting data from the output files ending in \_sum.lis and \_tab.lis has been processed and saved as fluka\_data.json. The Average Uncertainty and Total Primaries Run (nps) have been extracted. \\ \hline

\textbf{Check Average Uncertainty} & 
I will now check if the Average Uncertainty is less than 10\%. If it is, I will proceed to Step 8. Otherwise, I will proceed to Step 6. \newline
Let’s check the Average Uncertainty from the fluka\_data.json file. \newline
Here is the content of the fluka\_data.json file: \newline
\ttfamily \{
\newline
\textbf{‘output\_fort\_46\_tab.lis’}: \newline
\textbf{‘Average Uncertainty’: 8.5}, \newline
\textbf{‘output\_fort\_46\_sum.lis’}: \newline
\textbf{‘Total Primaries Run’: 1600000}
\newline
\} \normalfont \newline
The Average Uncertainty is 8.5\%, which is less than 10\%. Therefore, we can proceed to Step 8. \\ \hline

\textbf{Step 8: Plot Data} & 
I will now plot the data from the fluka\_data.json file with the following specifications: plot\_error\_bars=False, plot\_blocks=False, log\_scale=False, semilogx=True, semilogy=False. \\ \hline

\end{longtable}
\end{scriptsize}

\begin{figure}[H]
    \centering
    \begin{subfigure}[!t]{0.32\linewidth} 
        \centering
        \includegraphics[width=\linewidth]{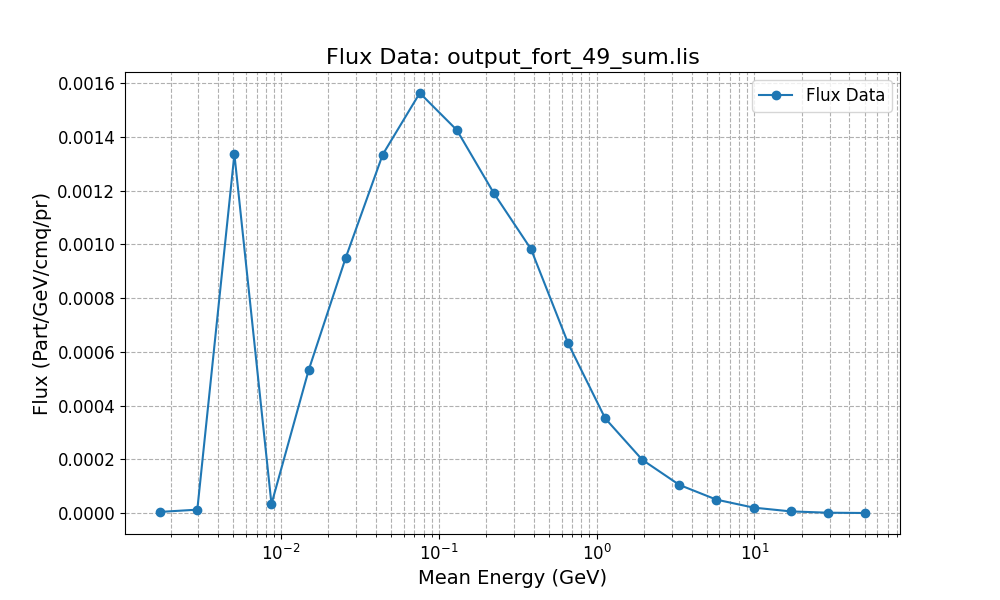}
        \caption{}
        \label{AppD_figA}
    \end{subfigure}
    \hfill
    \begin{subfigure}[!t]{0.32\linewidth} 
        \centering
        \includegraphics[width=\linewidth]{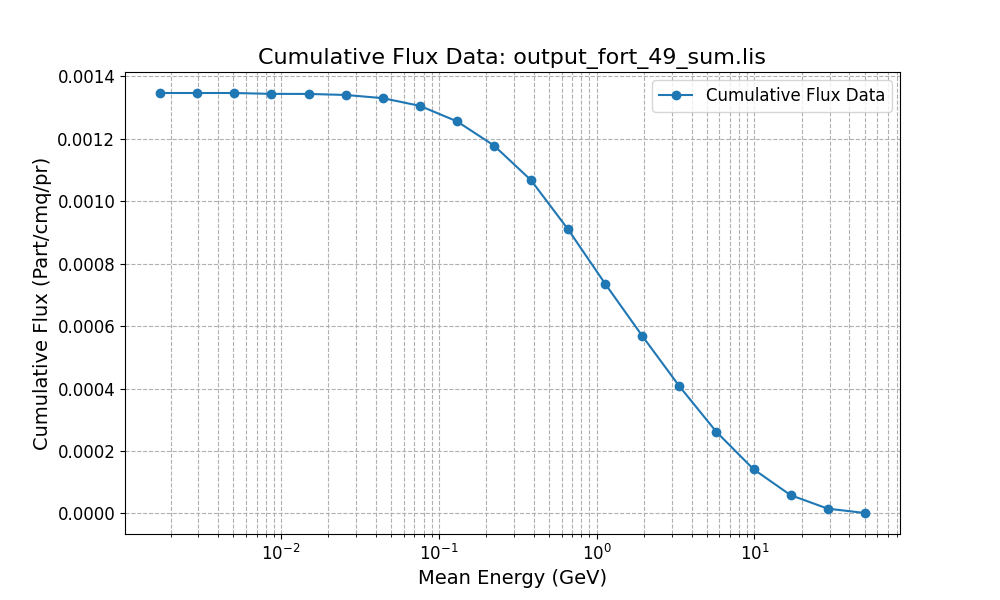}
        \caption{}
        \label{AppD_figB}
    \end{subfigure}
    \hfill
    \begin{subfigure}[!t]{0.32\linewidth} 
        \centering
        \includegraphics[width=\linewidth]{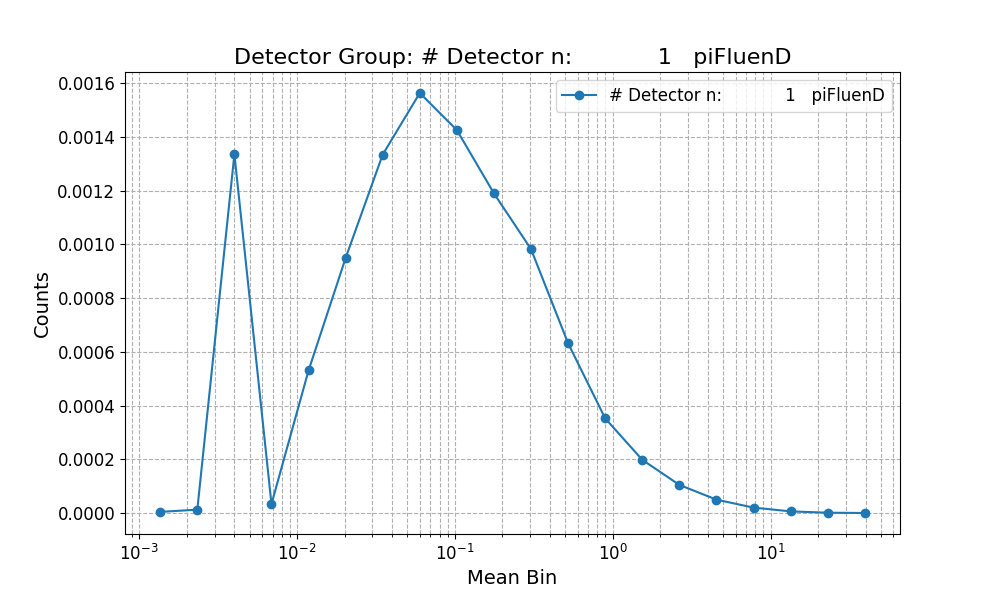}
        \caption{}
        \label{AppD_figC}
    \end{subfigure}
    
    \caption{Plots generated during the multi agent workflow; (a)- Spectral flux extracted from the \textbf{“\_sum.lis”} output file; (b)- cumulative flux also extracted from the \textbf{“\_sum.lis”} output file; (c)- Spectral flux extracted from the bin-wise \textbf{“\_tab.lis”} output file. Notice that this is identical to the \textbf{“\_sum.lis”} file data and was done for verification purposes.}
    \label{figD10:three-images}
\end{figure}

\section*{Data Availability}
The source code for AutoFLUKA will be made publicly available on GitHub after the official publication of this manuscript. The repository will include the full code, models, and instructions for use.

\section*{Acknowledgements}

The authors acknowledge Texas A\&M University’s High-Performance Computing (TAMU-HPRC) Department for providing access to their computing resources where most of the FLUKA simulations were executed.





\bibliographystyle{elsarticle-harv}
\bibliography{ref}  

\end{document}